\documentclass{article}
\usepackage[preprint]{neurips_2026}
\usepackage{fontspec}
\usepackage{unicode-math}
\usepackage{hyperref}
\hypersetup{colorlinks=true,linkcolor=blue,citecolor=blue,urlcolor=blue}
\usepackage{url}
\usepackage{booktabs}
\usepackage{amsmath}
\usepackage{graphicx}
\usepackage{longtable,array,calc}
\usepackage{xcolor}
\usepackage{microtype}
\usepackage{newunicodechar}
\newunicodechar{ℝ}{\ensuremath{\BbbR}}
\newunicodechar{∈}{\ensuremath{\in}}
\newunicodechar{⊙}{\ensuremath{\odot}}
\newunicodechar{𝒩}{\ensuremath{\mathcal{N}}}
\makeatletter
\newsavebox\pandoc@box
\newcommand*\pandocbounded[1]{%
  \sbox\pandoc@box{#1}%
  \Gscale@div\@tempa{\textheight}{\dimexpr\ht\pandoc@box+\dp\pandoc@box\relax}%
  \Gscale@div\@tempb{\linewidth}{\wd\pandoc@box}%
  \ifdim\@tempb\p@<\@tempa\p@\let\@tempa\@tempb\fi
  \ifdim\@tempa\p@<\p@\scalebox{\@tempa}{\usebox\pandoc@box}%
  \else\usebox{\pandoc@box}%
  \fi}
\def\fps@figure{htbp}
\makeatother
\providecommand{\tightlist}{\setlength{\itemsep}{0pt}\setlength{\parskip}{0pt}}
\title{Null-model treatment of the sensory-motor boundary changes an
evolutionary connectome comparison}
\author{Gyujeong Park\\
IonLabs\\
ORCID 0009-0006-2989-728X\\
\texttt{ionlabs2025@gmail.com}}

\begin{document}
\maketitle
\begin{abstract}
Randomised copies of a connectome are the usual baseline for asking
whether measured wiring matters, and the answer depends on what the
randomisation preserves. We evolved embodied foraging agents whose
brains are a compressed adult \emph{Drosophila} connectome (FlyWire
v783; 512 cell-type groups and 1,000 Kenyon cells) alongside agents
built on randomised wiring, in pre-registered experiments with ten
seeds, four ecologies and 600 generations. Two standard randomisations,
a column shuffle and degree-preserving edge swaps, route 10.6 to 10.7 \%
of olfactory output directly onto descending motor groups, against 0.012
\% in the connectome. On the registered primary endpoint, fitness
averaged over the run, no difference was detected; at the last
common-garden probe the connectome was behind both controls (−0.22 and
−0.20 fitness units on seed means). Against controls that keep every
sensory-output and motor-input edge and rewire only the interior, the
seed-mean difference lay within a ±0.10 equivalence bound (+0.002 and
−0.074, unchanged under a calibration that also matches activity
spread), although per ecology the interior column shuffle was ahead by
0.26 in one of four ecologies at ten seeds, a lead that ten further
pre-registered seeds did not replicate. Rewiring the connectome so that
it acquires the shortcut raised its fitness by 0.44 (10 of 10 seeds) and
its dependence on olfaction from 0.15 to 0.99; graded doses raised both
in step; at comparable swap counts the full dose was ahead of an
interior-only sham by 0.53 (10 of 10 seeds); and a sham that rewired the
same boundary edges without creating shortcuts matched the connectome
(+0.007) while the full dose was ahead of it by 0.60. What a null
preserves at the sensory-motor boundary can decide an evolutionary
connectome comparison, and sensory-to-motor path statistics belong next
to the degree statistics a null is said to preserve.
\end{abstract}

\section{Introduction}\label{sec-intro}

Whole-brain wiring diagrams of the adult fly are public {[}Dorkenwald et
al., 2024; Schlegel et al., 2024{]}, and models constrained by them
reproduce feeding circuits {[}Shiu et al., 2024{]}, visual responses
{[}Lappalainen et al., 2024{]}, head-direction dynamics {[}Duan et al.,
2025{]} and locomotion {[}Vaxenburg et al., 2025{]}. Whether the
measured wiring is better than chance is usually asked by comparing it
with a randomised network that preserves some of its statistics. The
verdict of such a comparison is a statement about the properties the
randomisation changed {[}Váša \& Mišić, 2022{]}, so the choice of null
decides what is being tested.

\begin{figure}
\centering
\includegraphics[width=1\linewidth,height=\textheight,keepaspectratio,alt={What the null preserves at the sensory-motor boundary, and what it changes. (a) In the compressed connectome olfactory input reaches descending (motor) groups through several synapses; the standard column-shuffle and degree-preserving nulls route about a tenth of olfactory output directly onto them; the boundary-preserving nulls keep every sensory-output and motor-input edge and replace about nine in ten interior edges. (b) Differences in evolved fitness at the last common-garden probe with 90 \% intervals: connectome minus each null on seed means over four ecologies (median), and the shortcut interventions over two ecologies (transplant: median; full dose minus sham and sham minus connectome: mean, as registered). The last two rows are the boundary-targeted sham (Section ), from separate runs with their own connectome island. The grey band is the ±0.10 equivalence bound.}]{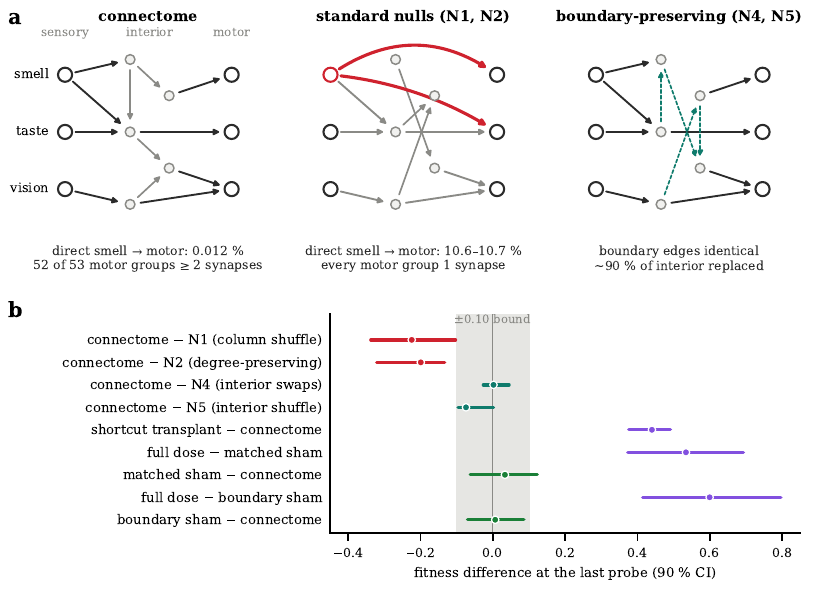}
\caption{What the null preserves at the sensory-motor boundary, and what
it changes. (a) In the compressed connectome olfactory input reaches
descending (motor) groups through several synapses; the standard
column-shuffle and degree-preserving nulls route about a tenth of
olfactory output directly onto them; the boundary-preserving nulls keep
every sensory-output and motor-input edge and replace about nine in ten
interior edges. (b) Differences in evolved fitness at the last
common-garden probe with 90 \% intervals: connectome minus each null on
seed means over four ecologies (median), and the shortcut interventions
over two ecologies (transplant: median; full dose minus sham and sham
minus connectome: mean, as registered). The last two rows are the
boundary-targeted sham (Section \ref{sec-bsham}), from separate runs
with their own connectome island. The grey band is the ±0.10 equivalence
bound.}\label{fig:overview}
\end{figure}

Our first pre-registered experiment found that randomised wiring
out-evolves a compressed fly connectome. Following that result up showed
that the two standard randomisations we had used move about a tenth of
olfactory receptor output directly onto descending motor neurons, a
class of connection that carries 0.012 \% of calibrated olfactory output
in the connectome, where 52 of 53 descending groups are two or more
synapses from olfactory input. This is not an error in those nulls; it
is what randomising the wiring between sensory and motor neurons does.
But in an embodied task where a short reactive route is useful, it
changes what the comparison measures. This paper reports what happens
when the null keeps the sensory-motor boundary fixed, and what happens
when the shortcut is moved into the connectome and out of the control
(Figure \ref{fig:overview}).

Contributions:

\begin{itemize}
\tightlist
\item
  \textbf{A measurement} of how two standard randomisations change the
  sensory-to-motor boundary of a connectome-derived agent: the share of
  olfactory output reaching descending groups directly rises from 0.012
  \% to 10.6 to 10.7 \%, and the median path length from olfactory
  receptors to descending groups falls from three synapses to one.
\item
  \textbf{A pre-registered comparison} against boundary-preserving
  controls, which keep every edge out of a sensory group and every edge
  into a motor group: the connectome's end-of-run disadvantage shrinks
  to within a ±0.10 equivalence bound on seed means, with the
  per-ecology heterogeneity and the null primary endpoint reported
  alongside.
\item
  \textbf{Pre-registered interventions} that move the shortcut:
  shortcut-creating rewiring raises the connectome's evolved fitness and
  transfers the controls' olfactory strategy to it; shortcut-removing
  rewiring costs a control its advantage where it had one; the effect
  rises with dose; an interior-only sham with a comparable number of
  swaps does not reproduce it; and neither does a sham that rewires the
  same boundary edges without creating shortcuts.
\item
  \textbf{Open protocols, runs and a dated research log}, including a
  retraction and a correction made after pre-registration, so that each
  result can be traced to the protocol that preceded it.
\end{itemize}

\section{Related work}\label{sec-related}

\textbf{Connectome-constrained models.} Fly connectome models are
usually evaluated on how well they reproduce measured activity or
behaviour {[}Shiu et al., 2024; Lappalainen et al., 2024; Duan et al.,
2025{]}. We instead ask whether measured wiring is a better starting
point for evolution than randomised wiring, a question closer to Worrell
et al.~{[}2017{]}, who found faster sensor-to-effector spreading in the
fly connectome than in degree-preserving nulls.

\textbf{Null models for connectomes.} A null keeps some properties and
randomises the rest, and constraining it to the property under test is
standard advice {[}Váša \& Mišić, 2022{]}; degree-preserving rewiring
misses spatially constrained structure {[}Salova \& Kovács, 2025{]}.
Dhiman {[}2026{]} found that a fly visual-system model's advantage over
degree-preserving nulls largely disappears under a shared random
initialisation, a verdict change through initialisation and degree
effects rather than path structure. Cho {[}2026{]} released a
behavioural benchmark for whole-connectome simulations with shuffled and
degree-preserving controls. We apply the constrain-the-null advice to
the sensory-motor boundary and show a case where it changes the outcome.

\textbf{Architecture, randomisation and measurement in machine
learning.} Architecture alone can encode useful behaviour {[}Gaier \&
Ha, 2019{]}, and which structure a randomised baseline keeps determines
what a comparison can attribute to it {[}Zhou et al., 2019{]}.
Randomisation tests are only informative about the properties they
destroy {[}Adebayo et al., 2018{]}, and an apparent effect can come from
the choice of measurement rather than from the system {[}Schaeffer et
al., 2023{]}. With ten seeds per condition we follow the recommendation
to report interval estimates over runs rather than point estimates
{[}Agarwal et al., 2021{]}.

\section{Methods}\label{sec-methods}

All quantities are dimensionless simulation units. Full specifications
are in Appendix \ref{app-model} (brain, dynamics, world, evolution) and
Appendix \ref{app-nulls} (controls).

\subsection{Agents}\label{sec-agents}

From FlyWire v783 we drop optic-lobe, visual-centrifugal and
photoreceptor classes, group the remaining neurons by cell type ×
hemisphere and keep K = 512 groups, starting with a mandatory set of
olfactory, gustatory, mushroom-body, dopaminergic, looming and named
descending groups and filling the rest by a sensory-to-descending path
score. Signed weights come from synapse counts and predicted
neurotransmitters {[}Eckstein et al., 2024{]}; the matrix has 14,989
non-zero entries, and sensory rows are zeroed so that sensors act as
transducers. Group rates follow leaky rectified-tanh dynamics. 1,000
Kenyon cells keep their measured projection-neuron inputs, with APL-like
inhibition holding 7 to 9 \% active, and a dopamine-gated plastic
Kenyon-cell-to-output matrix resets at birth. Forward speed, turning and
eating are read out from 24 bilateral descending-neuron types,
initialised from named descending neurons (e.g.~DNa01/DNa02 for
turning), with a per-brain innate feeding reflex. The compression keeps
named sensory and motor pathways on purpose, so the conditions compared
share this boundary and differ in the wiring between its two sides.

\subsection{World, ecologies and evolution}\label{sec-world}

A unit square holds 24 odor-marked patches (nutritious, potentially
toxic, neutral), with the odor-to-role assignment redrawn every lifetime
of 400 steps and an energy budget that forces foraging. Four ecologies
cross a predator that is stationary (P0) or pursuing (P1) with
potentially toxic patches that are all nutritious (T0) or all toxic
(T1). Each wiring condition evolves as its own island of 512 agents for
600 generations (tournament selection, 4 \% elitism); mutations perturb
weights, group sizes, leaks, biases, gains, learning rates and
read-outs, and can add or prune edges.

\subsection{Controls and interventions}\label{sec-controls}

Controls are drawn per seed and shared across that seed's ecologies
(Table \ref{tbl:nulls}). N1 moves each presynaptic column's non-zero
weights to random non-sensory rows; N2 performs Maslov--Sneppen target
swaps {[}Maslov \& Sneppen, 2002{]} with 10·\textbar E\textbar{}
attempts; N4 and N5 apply the same operations to \emph{interior} edges
only, whose source is not sensory and whose target is not a descending
group. The engine aborts unless N4 and N5 boundary edges are
bit-identical to the connectome; they replace about nine in ten interior
edges (interior overlap 0.111 to 0.121 for N4 and 0.089 to 0.095 for N5
over the 40 constructions of the corrected grid). The Kenyon-cell input
matrix is randomised by degree-preserving swaps in N2, N4 and N5 and by
a column shuffle in N1.

\begingroup\small

\begin{longtable}[]{@{}
  >{\raggedright\arraybackslash}p{(\linewidth - 4\tabcolsep) * \real{0.0714}}
  >{\raggedright\arraybackslash}p{(\linewidth - 4\tabcolsep) * \real{0.2857}}
  >{\raggedright\arraybackslash}p{(\linewidth - 4\tabcolsep) * \real{0.6429}}@{}}
\caption{Randomised controls. All keep edge counts and Dale
signs.}\label{tbl:nulls}\tabularnewline
\toprule\noalign{}
\begin{minipage}[b]{\linewidth}\raggedright
code
\end{minipage} & \begin{minipage}[b]{\linewidth}\raggedright
operation
\end{minipage} & \begin{minipage}[b]{\linewidth}\raggedright
additionally preserved
\end{minipage} \\
\midrule\noalign{}
\endfirsthead
\toprule\noalign{}
\begin{minipage}[b]{\linewidth}\raggedright
code
\end{minipage} & \begin{minipage}[b]{\linewidth}\raggedright
operation
\end{minipage} & \begin{minipage}[b]{\linewidth}\raggedright
additionally preserved
\end{minipage} \\
\midrule\noalign{}
\endhead
\bottomrule\noalign{}
\endlastfoot
N1 & column shuffle & out-degree, per-source weight multiset \\
N2 & degree-preserving target swaps & in- and out-degree \\
N4 & N2 on interior edges only & as N2, plus every sensory-output and
motor-input edge \\
N5 & N1 on interior edges only & as N1, plus every sensory-output and
motor-input edge \\
\end{longtable}

\endgroup

Three interventions move the shortcut directly. \textbf{AS} adds it to
the connectome: pairs of edges, one from an olfactory receptor group to
a non-motor target and one from a non-sensory source to a descending
group, exchange targets, which keeps every degree and each weight with
its source, until the olfactory shortcut share (the fraction of
olfactory output weight landing directly on descending groups) reaches
that of N2 (0.106). \textbf{N2R} runs the same operation in reverse on
N2 until its share is 0. A \textbf{dose} series raises the connectome's
share to at least 1, 3, 5 and 10 \%, and a \textbf{sham} performs 101
degree-preserving swaps on edge pairs whose source is not an olfactory
receptor and whose target is not a descending group, leaving the
shortcut share at 0.0001. The sham does not touch the boundary edges
that the dose rewires; §\ref{sec-dose} discusses what it therefore does
and does not control.

\subsection{Calibration}\label{sec-calibration}

Every network is row-normalised and scaled by a global and a Kenyon-cell
gain found by bisection so that mean group activity is 0.15 and mean
Kenyon-cell activity 0.05 over a fixed stimulus set. This matches mean
activity but not its spread: the connectome's median between-group
activity SD is 0.163 with 46 silent groups of 512, against medians of
0.080 to 0.087 and 8 to 19 silent groups for the controls (the three
standard nulls of the first grid, Appendix \ref{app-first}, and N4 and
N5 in the corrected grid). Row normalisation also means that a boundary
edge bit-identical before calibration can carry a slightly different
share of weight after it; all path statistics below are computed on the
calibrated matrices the runs used. A pre-registered re-run of the
corrected grid under a calibration that also matches the spread of
activity (between-group SD 0.118 and 3 to 7 silent groups in every
condition) returned the same verdicts (Section \ref{sec-equivalence},
Appendix \ref{app-grid}).

\subsection{Evaluation and statistics}\label{sec-stats}

Every 50 generations the 16 agents with the highest training fitness per
island are evaluated for 8 further lifetimes in a common garden: the
random-number generator is reset before the batch so that all conditions
and ablations face the same generated worlds. The registered
\textbf{primary endpoint} is this fitness averaged over generations 0 to
550 (trapezoid rule); the registered \textbf{secondary endpoint} is the
fitness at the last probe, generation 550. Olfaction, vision and
plasticity are ablated in parallel copies of the batch.

Seeds are the unit of replication. Paired differences (connectome minus
control) are tested with two-sided Wilcoxon signed-rank tests,
Holm-corrected across controls, and summarised by medians with
percentile bootstrap intervals over seeds (10,000 resamples). Runs of
one seed share their controls across ecologies, so cross-ecology
statements average each seed's ecologies first and test the ten seed
means. For absence of a difference we require equivalence: the 90 \%
interval of the median paired difference must lie inside ±δ with δ =
0.10. The bound was set after the first grid, when the shortcut effect
(at least 0.20) was known, and before any equivalence interval was
computed; Appendix \ref{app-grid} reports how each verdict depends on δ.

\subsection{Pre-registration and a correction}\label{sec-prereg}

Each experiment below was run under a protocol frozen before its data
(listed in Appendix \ref{app-prereg}), with engine hashes recorded. The
first grid used a heading-noise term with a non-zero mean that no
network could offset; it was found through a question raised in
AI-assisted review of a draft and corrected, and every later experiment
uses zero-mean noise. The first grid is reported because it motivated
the rest; its generation-0 result is not used (Appendix
\ref{app-prereg}).

\section{Results}\label{sec-results}

\subsection{Standard controls carry sensory-to-motor shortcuts and
out-evolve the connectome}\label{sec-shortcuts}

In the connectome, exactly two of 14,989 edges run from an olfactory
receptor group to a descending group, carrying 0.012 \% of calibrated
olfactory output; one descending group is one synapse from olfactory
input, 15 are two, 36 are three and one is four. A column shuffle or
degree-preserving swaps put 10.6 to 10.7 \% of olfactory output directly
onto descending groups and make every descending group reachable in one
synapse (Table \ref{tbl:paths}). The boundary-preserving controls keep
the connectome's structure; their small differences in the direct shares
come from calibration. For vision, whose routes are already short,
randomisation works the other way and lowers the direct share.

\begingroup\small

\begin{longtable}[]{@{}
  >{\raggedright\arraybackslash}p{(\linewidth - 8\tabcolsep) * \real{0.3538}}
  >{\raggedright\arraybackslash}p{(\linewidth - 8\tabcolsep) * \real{0.1538}}
  >{\raggedright\arraybackslash}p{(\linewidth - 8\tabcolsep) * \real{0.1538}}
  >{\raggedright\arraybackslash}p{(\linewidth - 8\tabcolsep) * \real{0.1692}}
  >{\raggedright\arraybackslash}p{(\linewidth - 8\tabcolsep) * \real{0.1692}}@{}}
\caption{Sensory-to-motor structure of the calibrated networks (medians
over 10 seeds). ``Direct'': share of that sense's output weight landing
on descending groups; ``hops'': median shortest path from that sense to
a descending group.}\label{tbl:paths}\tabularnewline
\toprule\noalign{}
\begin{minipage}[b]{\linewidth}\raggedright
condition
\end{minipage} & \begin{minipage}[b]{\linewidth}\raggedright
smell: direct
\end{minipage} & \begin{minipage}[b]{\linewidth}\raggedright
smell: hops
\end{minipage} & \begin{minipage}[b]{\linewidth}\raggedright
vision: direct
\end{minipage} & \begin{minipage}[b]{\linewidth}\raggedright
taste: direct
\end{minipage} \\
\midrule\noalign{}
\endfirsthead
\toprule\noalign{}
\begin{minipage}[b]{\linewidth}\raggedright
condition
\end{minipage} & \begin{minipage}[b]{\linewidth}\raggedright
smell: direct
\end{minipage} & \begin{minipage}[b]{\linewidth}\raggedright
smell: hops
\end{minipage} & \begin{minipage}[b]{\linewidth}\raggedright
vision: direct
\end{minipage} & \begin{minipage}[b]{\linewidth}\raggedright
taste: direct
\end{minipage} \\
\midrule\noalign{}
\endhead
\bottomrule\noalign{}
\endlastfoot
connectome & 0.00012 & 3 & 0.381 & 0.121 \\
N1 column shuffle & 0.107 & 1 & 0.108 & 0.113 \\
N2 degree-preserving & 0.106 & 1 & 0.128 & 0.093 \\
N4 interior swaps & 0.00010 & 3 & 0.340 & 0.125 \\
N5 interior shuffle & 0.00011 & 3 & 0.369 & 0.123 \\
\end{longtable}

\endgroup

In the first pre-registered grid (4 ecologies × 10 seeds, with the
uncorrected noise term), connectome populations were overtaken: at the
last probe the column shuffle was ahead in all ten seeds on seed means
(p\_holm = 0.0059) and the degree-preserving control in eight (p\_holm =
0.0195), while connectome populations depended far less on olfaction
(Appendix \ref{app-first}).

\subsection{Against boundary-preserving controls the seed-mean
difference lies within the equivalence bound}\label{sec-equivalence}

We re-ran the grid with the noise corrected and with N4 and N5 added (40
runs, 5 conditions each). On the registered primary endpoint no
difference was detected against any control (seed-mean medians −0.03 for
N1, −0.02 for N2, +0.03 for N4 and −0.01 for N5, none significant); the
effects below emerge late in evolution and are measured on the
registered secondary endpoint. At the last probe the connectome was
behind the shortcut-carrying controls (p\_holm = 0.029 for N1 and 0.016
for N2 on seed means), and against the boundary-preserving controls the
seed-mean difference lay inside the ±0.10 bound: +0.002 for N4 and
−0.074 for N5 (Table \ref{tbl:grid}). The N4 verdict holds for any bound
down to 0.05; the N5 verdict holds at 0.10 but not at 0.075 and is
marginal.

\begingroup\footnotesize

\begin{longtable}[]{@{}
  >{\raggedright\arraybackslash}p{(\linewidth - 10\tabcolsep) * \real{0.1765}}
  >{\raggedright\arraybackslash}p{(\linewidth - 10\tabcolsep) * \real{0.1647}}
  >{\raggedright\arraybackslash}p{(\linewidth - 10\tabcolsep) * \real{0.1647}}
  >{\raggedright\arraybackslash}p{(\linewidth - 10\tabcolsep) * \real{0.1647}}
  >{\raggedright\arraybackslash}p{(\linewidth - 10\tabcolsep) * \real{0.1647}}
  >{\raggedright\arraybackslash}p{(\linewidth - 10\tabcolsep) * \real{0.1647}}@{}}
\caption{Corrected grid, last probe (generation 550), connectome minus
control: median over 10 seeds, with the 90 \% interval below.
Equivalence is assessed only on seed means across
ecologies.}\label{tbl:grid}\tabularnewline
\toprule\noalign{}
\begin{minipage}[b]{\linewidth}\raggedright
control
\end{minipage} & \begin{minipage}[b]{\linewidth}\raggedright
seed means
\end{minipage} & \begin{minipage}[b]{\linewidth}\raggedright
P0T0
\end{minipage} & \begin{minipage}[b]{\linewidth}\raggedright
P1T0
\end{minipage} & \begin{minipage}[b]{\linewidth}\raggedright
P0T1
\end{minipage} & \begin{minipage}[b]{\linewidth}\raggedright
P1T1
\end{minipage} \\
\midrule\noalign{}
\endfirsthead
\toprule\noalign{}
\begin{minipage}[b]{\linewidth}\raggedright
control
\end{minipage} & \begin{minipage}[b]{\linewidth}\raggedright
seed means
\end{minipage} & \begin{minipage}[b]{\linewidth}\raggedright
P0T0
\end{minipage} & \begin{minipage}[b]{\linewidth}\raggedright
P1T0
\end{minipage} & \begin{minipage}[b]{\linewidth}\raggedright
P0T1
\end{minipage} & \begin{minipage}[b]{\linewidth}\raggedright
P1T1
\end{minipage} \\
\midrule\noalign{}
\endhead
\bottomrule\noalign{}
\endlastfoot
N1 (shortcut) & −0.224 & −0.621 & −0.312 & −0.159 & +0.054 \\
& {[}−0.34, −0.10{]} & {[}−0.66, −0.47{]} & {[}−0.44, −0.22{]} &
{[}−0.21, +0.01{]} & {[}−0.05, +0.13{]} \\
N2 (shortcut) & −0.199 & −0.403 & −0.338 & −0.017 & +0.015 \\
& {[}−0.32, −0.13{]} & {[}−0.42, −0.31{]} & {[}−0.53, −0.07{]} &
{[}−0.33, +0.01{]} & {[}−0.13, +0.14{]} \\
N4 (boundary) & +0.002 & −0.064 & +0.021 & −0.031 & +0.120 \\
& {[}−0.026, +0.045{]} & {[}−0.26, +0.00{]} & {[}−0.03, +0.04{]} &
{[}−0.12, +0.09{]} & {[}−0.06, +0.28{]} \\
N5 (boundary) & −0.074 & −0.264 & +0.067 & −0.015 & +0.070 \\
& {[}−0.096, +0.002{]} & {[}−0.37, −0.18{]} & {[}−0.07, +0.09{]} &
{[}−0.10, +0.06{]} & {[}−0.01, +0.21{]} \\
\end{longtable}

\endgroup

The seed-mean equivalence does not hold ecology by ecology, and we do
not claim it there. In the safe-food ecology (P0T0) the interior column
shuffle N5 is ahead of the connectome by 0.26 (p\_holm = 0.029), more
than twice the bound; in the predator-plus-toxin ecology the connectome
is ahead of N4 by 0.12 with an interval spanning zero. We then ran seeds
10 to 19 in the two T0 ecologies under rules fixed before those runs
(Appendix \ref{app-grid}). N5's lead in P0T0 did not replicate: in the
new seeds the connectome (A) minus N5 was +0.031 (90 \% CI −0.068 to
+0.152), and over all 20 seeds −0.104 (−0.208 to +0.016, p\_holm =
0.27). At n = 20 the per-ecology intervals for N4 and N5 lie inside
±0.10 in P1T0 (+0.003 and +0.056) but not in P0T0 (N4 −0.011, −0.114 to
+0.080). N1 stays ahead in both ecologies (−0.491 and −0.333, p\_holm ≤
0.004); N2 stays ahead in P1T0 (−0.203, p\_holm = 0.046), but its P0T0
lead shrank to −0.214 (−0.375 to +0.042, p\_holm = 0.25), and in seeds
10 to 19 alone the connectome was ahead of N2 there (+0.100). The larger
sample therefore weakens both the one per-ecology lead of a
boundary-preserving null and part of the standard nulls' per-ecology
advantage; the seed-mean contrast over four ecologies, which the
extension did not cover, is unchanged.

The sensory strategy follows the same split. Blocking olfaction costs
the connectome 0.03 against 0.68 for N1 and 0.64 for N2 (p\_holm =
0.008), and 0.10 for N4 and 0.17 for N5; the connectome's differences
from N4 and N5 (−0.07 and −0.11) are five- to ninefold smaller than
those from N1 and N2 and not significant after correction (p\_holm =
0.13 for both).

Because global calibration leaves the connectome with a wider spread of
activity than the controls, we re-ran the corrected grid (40 runs) under
a distribution-matched calibration that equalises mean, spread and
silent fraction across conditions, with verdict rules fixed beforehand.
Every run passed the manipulation check (control SD within 0.01 and
silent-group count within 10 of the connectome's). The verdicts held: on
seed means the boundary-preserving controls stayed inside ±0.10 (A − N4
+0.055, 90 \% CI +0.035 to +0.097; A − N5 −0.009, −0.076 to +0.046) and
the standard controls stayed ahead (A − N1 −0.274, −0.306 to −0.123; A −
N2 −0.146, −0.231 to −0.037). The calibration shifted the N2 contrast by
+0.101 (+0.016 to +0.133) and left the other three shifts undetermined,
and N5's lead in P0T0 did not appear under it either (−0.066, −0.320 to
+0.124; Appendix \ref{app-grid}).

\subsection{Moving the shortcut in and out}\label{sec-swap}

\begin{figure}
\centering
\includegraphics[width=1\linewidth,height=\textheight,keepaspectratio,alt={Intervention. Left and middle: fitness trajectories in the two ecologies (median and interquartile range over 10 seeds). Right: differences from the connectome at the last probe, seed means over both ecologies, with 90 \% intervals. Adding shortcut-creating rewiring (AS) reproduces the control's advantage; removing it from the degree-preserving control (N2R) removes the advantage where it was clear.}]{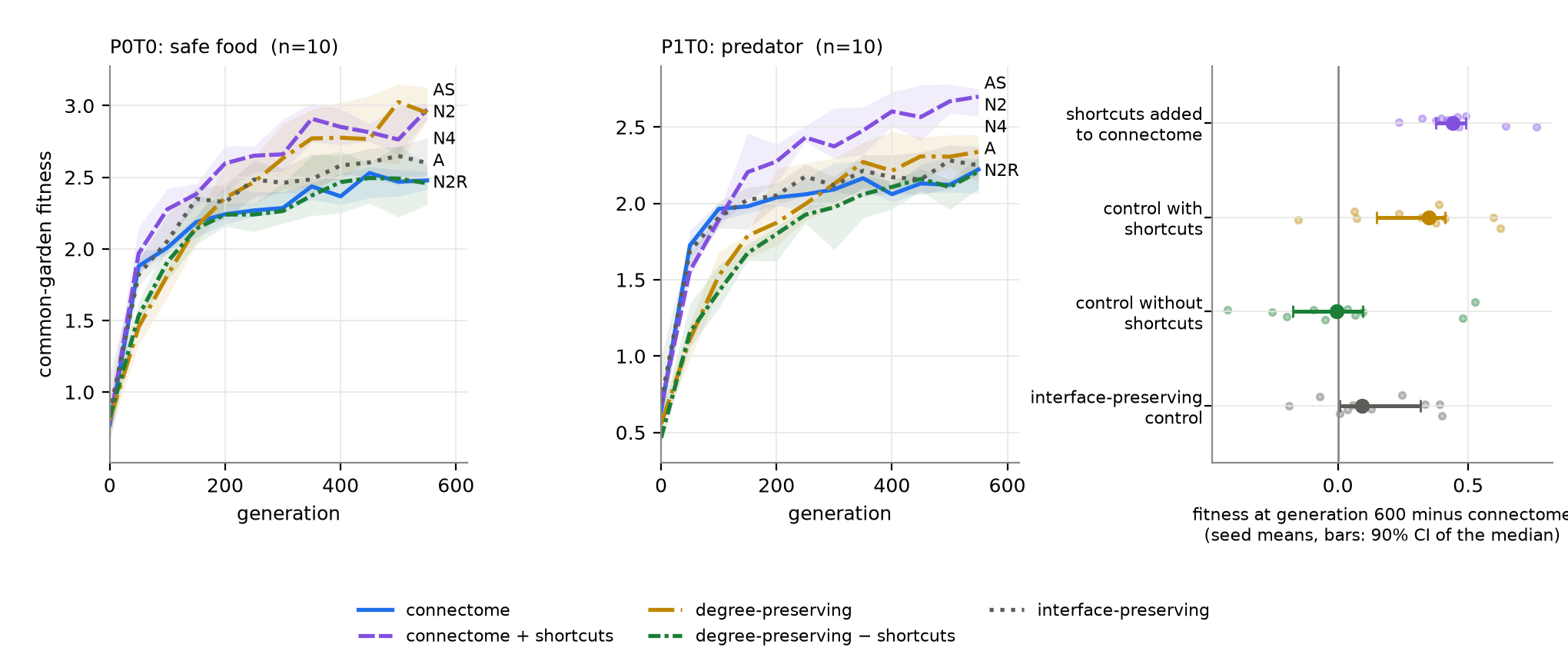}
\caption{Intervention. Left and middle: fitness trajectories in the two
ecologies (median and interquartile range over 10 seeds). Right:
differences from the connectome at the last probe, seed means over both
ecologies, with 90 \% intervals. Adding shortcut-creating rewiring (AS)
reproduces the control's advantage; removing it from the
degree-preserving control (N2R) removes the advantage where it was
clear.}\label{fig:swap}
\end{figure}

The pre-registered intervention (20 runs in the two ecologies with the
clearest overtaking) rewires the shortcut in both directions while
keeping degrees and Dale signs (Figure \ref{fig:swap}). Adding shortcuts
to the connectome raised its fitness at the last probe in every seed, by
+0.44 on seed means (10 of 10 seeds, p = 0.002; +0.445 in P0T0 and
+0.475 in P1T0), and its time-averaged fitness by +0.26 (10 of 10, p =
0.002). The effect grew over evolution: +0.04 at generation 0 (p = 0.32;
the interval is wider than the equivalence bound, so an initial effect
is not excluded) and +0.40 from generation 0 to the last probe (10 of 10
seeds, p = 0.002). The intervention also transferred the sensory
strategy, raising the cost of blocking olfaction from 0.15 to 0.99 (p =
0.002), close to the degree-preserving control's 0.84.

Removing the shortcut from the degree-preserving control cost it 0.43
fitness in the safe-food ecology (p = 0.004) and an estimated 0.00 in
the predator ecology (difference between ecologies −0.36, p = 0.027); in
neither ecology could the shortcut-free control be distinguished from
the connectome, and its olfactory dependence fell from 0.84 to 0.58 (p =
0.027). Both interventions also rewire edges that are not shortcuts, so
this experiment shows that \emph{shortcut-creating rewiring} is
sufficient for the advantage, not that the shortcut edges alone are.

\subsection{Graded doses and an interior-only sham}\label{sec-dose}

\begin{figure}
\centering
\includegraphics[width=1\linewidth,height=\textheight,keepaspectratio,alt={Graded shortcut transplant. Left and middle: fitness at the last probe against the share of olfactory output landing directly on descending groups, per ecology; grey lines are seeds, purple the mean with standard error, and the green square the interior-only sham (101 swaps) plotted at the connectome's own share. Right: fitness lost when olfaction is blocked, against the same axis.}]{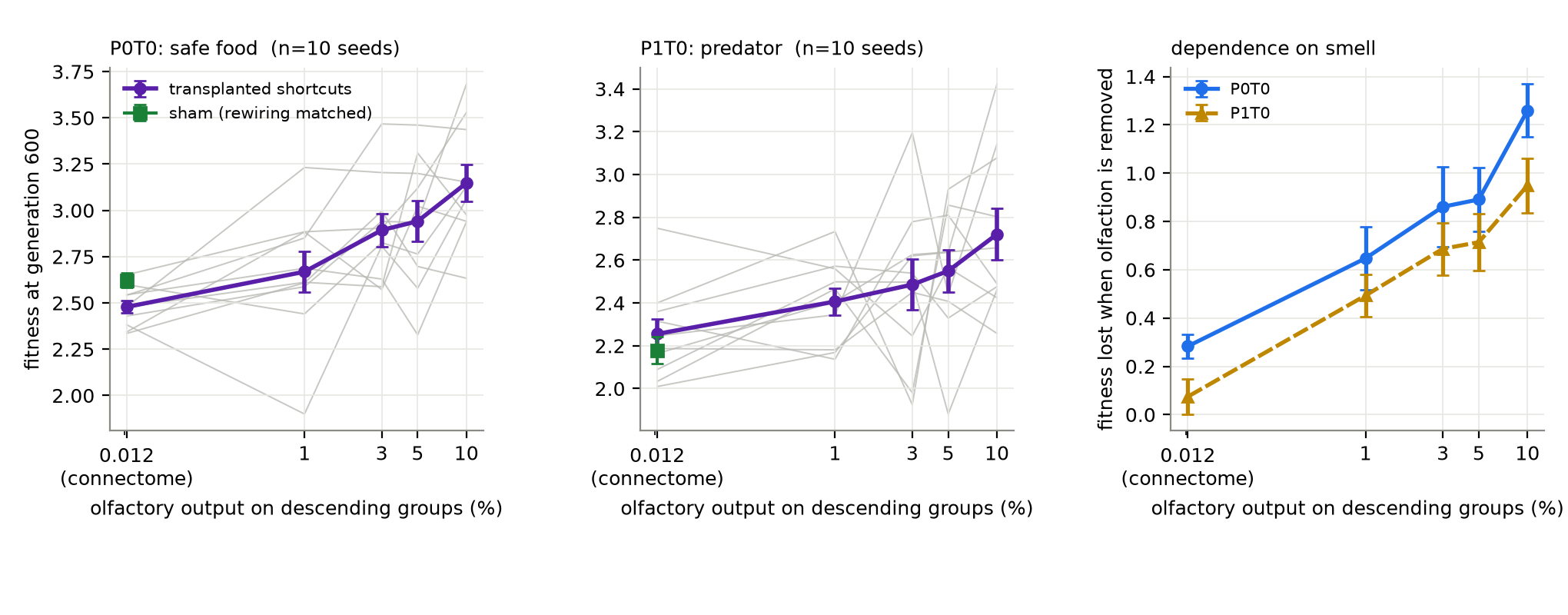}
\caption{Graded shortcut transplant. Left and middle: fitness at the
last probe against the share of olfactory output landing directly on
descending groups, per ecology; grey lines are seeds, purple the mean
with standard error, and the green square the interior-only sham (101
swaps) plotted at the connectome's own share. Right: fitness lost when
olfaction is blocked, against the same axis.}\label{fig:dose}
\end{figure}

To ask whether the effect follows the amount of shortcut, a
pre-registered dose series raised the connectome's
olfactory-to-descending share to at least 1, 3, 5 and 10 \% (2 ecologies
× 10 seeds). Fitness rose across doses, on seed means 2.37, 2.54, 2.69,
2.75 and 2.93 from the connectome to the 10 \% dose (Page's L, z = 4.68,
p = 1.4 × 10\textsuperscript{−6}; per-seed Spearman correlation positive
in 9 of 10 seeds), in both ecologies (p = 1.5 × 10\textsuperscript{−5}
and 0.004), and dependence on olfaction rose with it (p = 3.4 ×
10\textsuperscript{−5} and 5.6 × 10\textsuperscript{−7}; Figure
\ref{fig:dose}). At generation 0 no trend was detected across doses (p =
0.98).

The full dose took 68 to 129 swaps depending on the seed (mean 98.5),
comparable to the sham's 101, and was ahead of it by 0.534 on seed means
(90 \% CI +0.373 to +0.693, 10 of 10 seeds; +0.525 in the safe-food and
+0.543 in the predator ecology). It was ahead in all seven seeds in
which it used fewer swaps than the sham. The number of degree-preserving
swaps therefore does not account for the effect. What the sham does not
control is \emph{where} the swaps land: the dose necessarily rewires
olfactory-output and descending-input edges, while the sham avoids them,
so this comparison separates shortcut-creating rewiring from
interior-only rewiring, not the shortcut edges from other changes at the
boundary. Section \ref{sec-bsham} closes that gap with a sham that
rewires the same boundary edges. In this dose series shams were run only
at the full swap count; the boundary-targeted experiment pairs a sham
with every dose.

The sham itself was registered to fall inside the equivalence bound
against the connectome and did not: +0.034 on seed means (90 \% CI
−0.062 to +0.123), ahead by 0.144 in the safe-food ecology (interval
+0.072 to +0.222) and behind by 0.077 in the predator ecology (interval
spanning zero). Interior-only rewiring is therefore not demonstrably
inert, but its effect is small next to the dose's and does not keep its
sign across ecologies (Appendix \ref{app-dose}).

\subsection{A boundary-targeted sham separates the shortcut from
boundary rewiring}\label{sec-bsham}

A shortcut swap turns ORN→X and Y→DN into ORN→DN and Y→X. The
boundary-targeted sham BS\emph{k} rewires the same two boundary edges
without creating a shortcut: it takes a third, existing interior edge
Z→W and rotates the three targets, giving ORN→W, Y→X and Z→DN. The
olfactory weight lands in the interior instead of on a descending group,
and the descending group receives Z's weight instead. Each seed performs
as many rotations as its own AS\emph{k} performed swaps. Z→W is the one
of 64 random interior edges whose weight is closest in magnitude to
ORN→X (median ratio 0.92 to 1.05 across constructions). Degrees, signs
and the olfactory-to-descending share (0.0001) are unchanged, which the
engine checks. Nine islands (A; AS1, AS3, AS5, AS10; and their shams
BS1, BS3, BS5, BS10) evolved together in the two ecologies over ten
seeds, under a protocol frozen before the runs
(\texttt{PROTOCOL\_BSHAM.md}). The primary contrast is AS10 − BS10 at
the last probe, with a 90 \% bootstrap interval of the seed mean:
\emph{shortcut-specific} if the lower bound exceeds +0.10,
\emph{boundary rewiring suffices} if the interval lies inside ±0.10.

The verdict was shortcut-specific. The full dose was ahead of its
boundary-targeted sham by 0.600 (90 \% CI +0.414 to +0.797), and by
0.568 and 0.631 in the two ecologies. The registered robustness check
re-evaluated the generation-599 elites in four newly generated worlds
and gave the same category (+0.639, +0.490 to +0.792). The sham itself
was indistinguishable from the connectome (BS10 − A = +0.007, −0.070 to
+0.086, inside ±0.10). The gap between each dose and its matched sham
grew with dose (0.16, 0.28, 0.36 and 0.60 for 1, 3, 5 and 10 \%; Page's
L, z = 3.18, p = 7.4 × 10\textsuperscript{−4}). Blocking olfaction cost
the full dose 1.18 and its sham 0.35. The transplant's gain over the
connectome replicated (AS10 − A = +0.606, +0.435 to +0.798). Rewiring
the boundary edges the same number of times, without creating direct
olfactory-to-descending connections, therefore reproduced none of the
effect: it follows the shortcut edges themselves. Three runs failed with
GPU out-of-memory (one of them twice), one completed attempt was deleted
by a driver error and one attempt was stopped to test a memory fix;
every affected run was re-run from scratch with identical flags before
any outcome was inspected. Sixteen of the 20 runs used an engine copy
that differs only in forcing evaluation earlier to save memory and was
bit-identical to the original on the CPU (Appendix \ref{app-dose}).

\subsection{Further checks}\label{sec-checks}

Zeroing interior weights of evolved populations cost 18 to 23 \% of
fitness and permuting them 46 to 55 \%, in every condition, with no
detected difference between the connectome and any control (all p ≥
0.16). The evolved agents are therefore sensitive to their interior
wiring rather than running on the boundary alone; this is perturbation
sensitivity, not evidence that the task requires interior computation.
In the first grid, the fraction of one-step mutants beating their
parent's clones was similar for connectome and control genomes (9.6 \%
against 9.4 and 9.8 \% at generation 0), and connectome genomes lost
more to large mutations (Appendix \ref{app-first}). We detected no
odor-specific associative learning in any condition. In an earlier
version of the model (paper-v1, Appendix \ref{app-v1}), weights moved
over evolution by 0.5 to 3 \% of their initial spread while gains, leaks
and read-outs moved by about 6 to 11 initial standard deviations; we did
not repeat this measurement in the grids reported here (Appendix
\ref{app-checks}).

\section{Discussion}\label{sec-discussion}

In this model the connectome's standing depended on how the randomised
control treated connections between sensors and motor neurons. The two
standard nulls randomised the sensory-motor boundary along with
everything else, so the comparison they support is about the boundary as
much as about the interior wiring we meant to test. With the boundary
held fixed, the end-of-run seed-mean difference fell within ±0.10;
moving the shortcut into the connectome reproduced the controls'
advantage and olfactory strategy; and the advantage followed the dose. A
sham that rewired the same boundary edges the same number of times
without creating shortcuts left fitness where the connectome was, so the
effect is attributable to the direct olfactory-to-descending connections
rather than to rewiring at the boundary as such.

Which properties a null should preserve depends on the hypothesis. A
randomisation meant to test whether interior wiring matters should not
change which sensors reach which effectors in one step; one meant to
test the whole architecture may. Keeping every sensory-output and
motor-input edge, as N4 and N5 do, is one way to hold the boundary
fixed, and reporting calibrated sensory-to-motor path statistics next to
degree statistics makes the choice visible. The same concern applies
wherever a task rewards short reactive routes; whether it matters in
static benchmarks depends on their tasks, and the statistics are cheap
to report either way. Together with Dhiman {[}2026{]}, who found a
verdict change through initialisation and degree effects, this suggests
that which null was used, and what it preserved, matters more for
connectome comparisons than is usually acknowledged.

What we can say about the connectome itself is narrow. On seed means
over four ecologies we found no end-of-run advantage for measured
interior wiring over boundary-preserving controls, within ±0.10; that is
a bounded null for one estimand, not proof of no effect, and the one
per-ecology lead of the interior column shuffle did not replicate in ten
further seeds. Per-ecology estimates are also less stable than the seed
means: the degree-preserving control's lead in the safe-food ecology,
which motivated the intervention there, shrank to an interval spanning
zero at twenty seeds. Our agents are a compressed rate model in a
two-dimensional world, no condition evolved detectable associative
learning, and in an earlier version of the model evolution adapted gains
and read-outs far more than weights. A task that requires what fly
circuits are built for, such as learning from experience within a
lifetime, could give a different answer, and is the natural next test.

\textbf{Limitations.} One connectome, one compression and one family of
two-dimensional worlds; one mutation operator, which tunes far more than
it rewires; a toxin factor confounded with food abundance and a
stationary rather than absent predator; global calibration in the main
experiments (verdicts unchanged under distribution matching); a
common-garden evaluation that reuses one generated world stream per
probe; population samples rather than elites in the post hoc assays;
equivalence bounded at ±0.10 on seed means with ten seeds and only where
stated (twenty in two ecologies), with the bound set after the first
grid; interventions that rewire more than the shortcut edges (the
boundary-targeted sham controls this at the boundary, not everywhere),
an interior-only sham that failed its own registered prediction; and
boundary edges that mutation can add during evolution.

\section*{Data and code availability}\label{sec-data}
\addcontentsline{toc}{section}{Data and code availability}

Engines, analysis and figure scripts, frozen protocols, the dated
research log and run outputs are at
https://github.com/gyujeongion/flyconnectome-nulls; an archived copy
with the manuscripts is at https://doi.org/10.5281/zenodo.22871090. Raw
per-generation logs and stored genomes (about 2 GB) are not included;
the analysis outputs computed from them are. Re-running a seed does not
repeat a run bit for bit: after every common-garden probe, which runs
under a fixed random seed, the engines reseed the evolutionary random
stream from the clock, and the GPU scatter used in mutation does not fix
which write survives when two mutations hit the same weight. A re-run
reproduces the null constructions, calibration and generation 0 exactly
and later generations only in distribution; seeds are therefore
independent replicates, not reproducible trajectories. FlyWire v783
connectivity and annotations are public. Runs used one Apple M1 Ultra
with MLX (about 36 minutes per 5-condition, 600-generation run), except
the boundary-targeted sham runs, which used an Apple M2 Max (Appendix
\ref{app-bsham}).

\section*{Use of AI tools}\label{sec-ai}
\addcontentsline{toc}{section}{Use of AI tools}

The engines, analysis and figure scripts were written, and this
manuscript drafted, with an AI coding assistant (Claude Code,
Anthropic); two further language models critiqued drafts. Every number
is regenerated from stored run outputs by scripts in the repository, and
the author directed the study and is responsible for the text (Appendix
\ref{app-ai}).

\section*{Funding and conflicts of interest}\label{sec-funding}
\addcontentsline{toc}{section}{Funding and conflicts of interest}

This research received no external funding. The author declares no
conflicts of interest.

\section*{References}\label{references}
\addcontentsline{toc}{section}{References}

Adebayo, J., Gilmer, J., Muelly, M., Goodfellow, I., Hardt, M. \& Kim,
B. (2018) Sanity checks for saliency maps. \emph{Advances in Neural
Information Processing Systems} 31.

Agarwal, R., Schwarzer, M., Castro, P. S., Courville, A. \& Bellemare,
M. G. (2021) Deep reinforcement learning at the edge of the statistical
precipice. \emph{Advances in Neural Information Processing Systems} 34.

Cho, B. (2026) flybench: a behavioural benchmark for whole-connectome
simulations of \emph{Drosophila}, v0.1.0 (software).
https://github.com/brandoncho369/flybench

Dhiman, N. (2026) Topological sensitivity in connectome-constrained
neural networks. arXiv:2604.04033.

Dorkenwald, S. et al.~(2024) Neuronal wiring diagram of an adult brain.
\emph{Nature} 634, 124--138.

Duan, S., Dong, L. L. \& Fiete, I. (2025) From synapses to dynamics:
obtaining function from structure in a connectome constrained model of
the head direction circuit. \emph{Advances in Neural Information
Processing Systems} 38.

Eckstein, N. et al.~(2024) Neurotransmitter classification from electron
microscopy images. \emph{Cell} 187, 2574--2594.

Gaier, A. \& Ha, D. (2019) Weight agnostic neural networks.
\emph{Advances in Neural Information Processing Systems} 32.

Lappalainen, J. K. et al.~(2024) Connectome-constrained networks predict
neural activity across the fly visual system. \emph{Nature} 634,
1132--1140.

Maslov, S. \& Sneppen, K. (2002) Specificity and stability in topology
of protein networks. \emph{Science} 296, 910--913.

Salova, A. \& Kovács, I. A. (2025) Combined topological and spatial
constraints are required to capture the structure of neural connectomes.
\emph{Network Neuroscience} 9, 181--206.

Schaeffer, R., Miranda, B. \& Koyejo, S. (2023) Are emergent abilities
of large language models a mirage? \emph{Advances in Neural Information
Processing Systems} 36.

Schlegel, P. et al.~(2024) Whole-brain annotation and multi-connectome
cell typing of \emph{Drosophila}. \emph{Nature} 634, 139--152.

Shiu, P. K. et al.~(2024) A \emph{Drosophila} computational brain model
reveals sensorimotor processing. \emph{Nature} 634, 210--219.

Vaxenburg, R. et al.~(2025) Whole-body physics simulation of fruit fly
locomotion. \emph{Nature} 643, 1312--1320.

Váša, F. \& Mišić, B. (2022) Null models in network neuroscience.
\emph{Nature Reviews Neuroscience} 23, 493--504.

Worrell, J. C., Rumschlag, J., Betzel, R. F., Sporns, O. \& Mišić, B.
(2017) Optimized connectome architecture for sensory-motor integration.
\emph{Network Neuroscience} 1, 415--430.

Zhou, H., Lan, J., Liu, R. \& Yosinski, J. (2019) Deconstructing lottery
tickets: zeros, signs, and the supermask. \emph{Advances in Neural
Information Processing Systems} 32.

\appendix

\section{Model details}\label{app-model}

\subsection{Brain model}\label{brain-model}

\textbf{Compression.} From FlyWire v783 we drop optic-lobe,
visual-centrifugal and photoreceptor classes, group the remaining
neurons by cell type × hemisphere (mushroom-body output, dopaminergic
and MBIN types merged across hemispheres), and keep K = 512 groups. A
mandatory set is kept first: 16 olfactory receptor-neuron types
including the CO\textsubscript{2} channel ORN\_V; sugar and bitter
gustatory receptor neurons; every antennal-lobe projection-neuron group
receiving at least 5 \% of its input from kept ORNs plus the 24 most
ORN-driven local-neuron groups; mushroom-body output, dopaminergic and
modulatory neurons; APL and DPM; the looming detectors LC4 and LPLC2;
the descending neurons DNa01, DNa02, DNp09, MDN, DNp01, DNg11, DNg12\_a;
the 48 highest-scoring groups on a gustatory-to-dopaminergic path score;
PFL3 output targets and the path from them to DNa01/DNa02. Remaining
slots go to the highest scores on

score(i) = f(i)·b(i), f = Σ\_\{h=1..4\}(T\^{}h 1\_sensory)(i), b =
Σ\_\{h=1..4\}((T\textsuperscript{T})\^{}h 1\_descending)(i), T =
D\textsuperscript{−1} S\_abs,

with S\_abs the group-level absolute synapse-count matrix and D its row
sums. Weights are W\textsuperscript{0}\_ij = S\_signed,ij / (Σ\_k
S\_abs,ik + 1) with neurotransmitter-predicted signs (Eckstein et al.,
2024). Sensory rows are zeroed, so sensors act as transducers. The
resulting matrix has 14,989 non-zero entries.

The compression keeps named sensory and motor pathways on purpose, and
the read-outs start from named descending neurons, so the comparison is
between agent designs that share this connectome-derived boundary and
differ in the wiring between the two sides of it.

\textbf{Dynamics.} With group rates r ∈ ℝ\^{}K, size factors c = exp(s),
leak α = σ(a − 1), bias b, sensor vector S ∈ ℝ\textsuperscript{40} with
gains g\_in and fixed projection B\_in from sensor channels onto their
receptor groups, Kenyon-cell rates k, plastic KC→MBON matrix W\_km,
fixed KC→group matrix W\_ko and MBON one-hot map P\_MB:

x = W (r ⊙ c) + W\_ko k + P\_MB (W\_km k) + B\_in (S ⊙ g\_in) + b r ← (1
− α) ⊙ r + α ⊙ tanh(max(x, 0))

1,000 Kenyon cells keep their measured projection-neuron inputs W\_pk
(10.3 non-zero group inputs each). With u = g\_kc · W\_pk (r ⊙ c), k =
tanh(max(u − (mean(u) + λ·SD(u)), 0)) with heritable λ, an APL-like
inhibition keeping 7 to 9 \% of Kenyon cells active. Plasticity is
dopamine-gated: d = M\_DAN r\_DAN, d̄ ← 0.95 d̄ + 0.05 d, W\_km ←
clip(W\_km + η ⊙ (d − d̄) k\textsuperscript{T}, 0, 2), with heritable
per-MBON rates η and measured DAN→MBON connectivity. Plastic weights
reset at birth.

\textbf{Read-outs and actions.} For the 24 descending-neuron types
present on both sides, with rates D\_L and D\_R: forward =
R\textsubscript{0}·(D\_L + D\_R) + R\_M,0 r\_MBON + b\textsubscript{0};
turn = R\textsubscript{1}·((D\_L − D\_R) − m) with m ← 0.9 m + 0.1 (D\_L
− D\_R); eat = R\textsubscript{2}·(D\_L + D\_R) + R\_M,1 r\_MBON +
b\textsubscript{2}. R\textsubscript{0} and R\textsubscript{1} start from
named descending neurons (forward 2·DNp09 − 2·MDN + DNp01; turn 2·DNa01
+ 2·DNa02) scaled by 8 plus 𝒩(0, 0.3); R\textsubscript{2} is an innate
feeding reflex derived per brain from its own descending responses to
sugar, bitter and odor. Speed is v = v\_max σ(forward), reduced to 0.2
while eating; the agent eats when eat \textgreater{} 0; v\_max = 0.012,
ω\_max = 0.35. Heading follows

θ ← θ + ω\_max tanh(turn) + 0.05 (ξ − 0.5), ξ \textasciitilde{} U(0, 1).

The first grid (Appendix \ref{app-first}) used 0.05 ξ instead, which
adds a constant 0.025 rad per step; Appendix \ref{app-prereg} gives the
consequences. All later experiments use the zero-mean form above.

\subsection{World}\label{world}

A unit square with reflecting walls holds 24 patches: 10 nutritious, 8
potentially toxic, 6 neutral, each carrying one of 8 odors as a
15-dimensional glomerular vector, with the odor-to-role assignment
redrawn every lifetime. Antennae sit 0.05 from the body at ±0.6 rad;
odor concentration is Σ\_p
exp(−d\textsuperscript{2}/2σ\textsuperscript{2}) with σ = 0.06; predator
odor enters the CO\textsubscript{2} channel with σ\_p = 0.12; looming
input to LC4/LPLC2 is clip(0.04/d, 0, 1) on the predator's side when it
is in the frontal field. Eating radius 0.055; a nutritious bite gives
+0.04 energy, a toxic bite costs g\_bad = 0.25 and triggers a delayed
malaise signal reaching the bitter channel, and toxic food tastes sweet
on contact. Nutritious patches hold 12 bites, toxic ones 1; depleted
patches respawn elsewhere. Energy starts at 0.7 and falls by 0.0025 +
0.0008·v/v\_max + C\_brain per step, with C\_brain =
0.1·(0.0006·N\_eff/N\_ref + 0.0004·n\_edges/n\_edges,ref), where N\_eff
is the neuron count implied by the evolved group sizes, N\_ref = 13,718
the count of the initial brain, which is the 8,541 neurons of the 512
groups plus a fixed offset of 5,177 for the Kenyon-cell population,
which does not evolve, n\_edges the number of non-zero weights and
n\_edges,ref = 14,989; an agent that never eats dies before its lifetime
ends. Fitness is

fitness = 0.04·(nutritious bites) − g\_bad·(toxic bites) −
C\_brain·(steps alive) + 0.3·(alive at end) + 0.0005·(steps alive),

so the brain cost is charged twice by design, once through energy and
once directly.

\textbf{Ecologies.} Predator: stationary and harmless (P0), or pursuing
at 0.2·v\_max within 0.45 and killing within 0.035 (P1). In P0 the
predator stays put and still emits CO\textsubscript{2} and looming cues;
only capture is disabled. Toxin: the 8 potentially toxic patches are all
nutritious (T0, 18 nutritious patches) or all toxic (T1, 10 nutritious
patches), so the toxin factor also changes food abundance. The four
cells are P0T0, P1T0, P0T1, P1T1.

\subsection{Evolution}\label{evolution}

One island per wiring condition, 512 agents each, tournament selection
(k = 4), 4 \% elitism, 600 generations, no migration. Each child gets
3,000 random entries of W multiplied by exp(𝒩(0, 0.25)); 3 zero entries
turned into edges with Dale-correct sign and magnitude \textbar 𝒩(0,
0.05)\textbar, which may fall anywhere including the boundary; 60 random
entries pruned with probability 0.5; entries below
10\textsuperscript{−3} zeroed. Gaussian perturbations (probability, SD,
clip): group size (0.05, 0.15, {[}−2, 2{]}), leak (0.05, 0.3, {[}−4,
4{]}), bias (0.05, 0.1, {[}−2, 2{]}), learning rates (0.2, 0.03, {[}−1,
1{]}), input gains (0.3, 0.2, {[}0, 8{]}), Kenyon-cell gain (0.3, 0.1,
{[}−2, 2{]}), Kenyon-cell sparsity (0.3, 0.1, {[}0.5, 3{]}), descending
read-out (0.2, 0.5, {[}−12, 12{]}), MBON read-out (0.2, 0.5, {[}−10,
10{]}), read-out biases (0.3, 0.2, {[}−5, 5{]}).

Every 250 generations the run stores 8 genomes per island. An indexing
error makes these members of the population just after selection rather
than the elites; we call them population samples and use them only in
the post hoc assays of Appendix \ref{app-checks}. Their fitness relative
to the top 16 of the same generation is 0.98 (connectome), 0.93 (N1) and
0.98 (N2), with paired tests against the connectome giving p = 0.14 and
0.65; this does not prove the sampling is unbiased, and the assays that
rest on these genomes carry that caveat.

\section{Controls, interventions and calibration}\label{app-nulls}

\subsection{Definitions}\label{definitions}

Controls are drawn fresh per seed from one random stream, in the order
the conditions are listed, and are used in all ecologies of that seed.
W\_pk is randomised by the same rule; W\_ko and initial KC→MBON weights
are shared. The first three are in common use. The last two apply the
general practice of constraining a null to the property under test (Váša
\& Mišić, 2022) to the sensory-motor boundary; we did not find this
particular constraint in earlier connectome comparisons, but it is an
application of that practice, not a new principle.

{
\begin{longtable}[]{@{}
  >{\raggedright\arraybackslash}p{(\linewidth - 4\tabcolsep) * \real{0.3333}}
  >{\raggedright\arraybackslash}p{(\linewidth - 4\tabcolsep) * \real{0.3333}}
  >{\raggedright\arraybackslash}p{(\linewidth - 4\tabcolsep) * \real{0.3333}}@{}}
\toprule\noalign{}
\begin{minipage}[b]{\linewidth}\raggedright
code
\end{minipage} & \begin{minipage}[b]{\linewidth}\raggedright
operation
\end{minipage} & \begin{minipage}[b]{\linewidth}\raggedright
preserved
\end{minipage} \\
\midrule\noalign{}
\endhead
\bottomrule\noalign{}
\endlastfoot
N1 & permute the row positions of each presynaptic column's non-zeros &
out-degree, Dale sign, per-source weight multiset \\
N2 & Maslov--Sneppen target swaps (Maslov \& Sneppen, 2002),
10·\textbar E\textbar{} attempts & in- and out-degree, weight--source
pairing, sign \\
N3 & permute entries within each (super-class × hemisphere) block pair &
connections between block pairs, block-level laterality \\
N4 & Maslov--Sneppen swaps restricted to interior edges & as N2, plus
every edge leaving a sensory group and every edge entering a descending
group \\
N5 & column shuffle restricted to interior edges & as N1, plus the same
boundary edges \\
\end{longtable}
}

Interior edges are those whose source is not a sensory group and whose
target is not a descending group. For N4 and N5 the engine checks at
construction that the boundary edges are bit-identical to the connectome
(maximum absolute difference 0) and aborts otherwise; edge counts match
the connectome exactly, and interior edge overlap with the connectome is
0.09 to 0.12 (N4 0.111 to 0.121, N5 0.089 to 0.095 over the 40
constructions of the corrected grid), so both replace roughly nine in
ten interior edges. Boundary preservation constrains the first and last
synapse of any sensory-to-motor route; it does not fix the lengths or
identities of the interior routes between them, which is what these
controls are meant to randomise.

Two further conditions manipulate the shortcut directly. \textbf{AS} is
the connectome with shortcuts added: pairs of edges, one from an
olfactory receptor group to a non-motor target and one from a
non-sensory source to a descending group, exchange targets, which
preserves every in- and out-degree and keeps each weight with its
source. Swaps continue until the olfactory shortcut share reaches 0.106,
the value of the degree-preserving control. \textbf{N2R} is the
degree-preserving control with the same operation run in reverse until
its shortcut share is 0. Both interventions rewire more than the
shortcut edges themselves, so they test shortcut-creating and
shortcut-removing rewiring rather than an isolated variable. The two
activity statistics we monitor are unchanged in AS (between-group SD
0.160 against 0.163, 44 silent groups against 48); other aspects of the
dynamics may differ.

The \textbf{olfactory shortcut share} is the fraction of the absolute
output weight of olfactory receptor groups that lands directly on
descending groups.

\subsection{Implementation}\label{implementation}

The randomisations are implemented in \texttt{evo\_fix.py}
(\texttt{shuffle\_control}, \texttt{null\_degswap},
\texttt{null\_interface\_degswap},
\texttt{null\_interface\_colshuffle}); the dose and sham operations in
\texttt{evo\_dose.py}. Precisely:

\begin{itemize}
\tightlist
\item
  \textbf{Sensory rows.} Sensory groups receive no input in the
  connectome (their rows are zero). N1 places each column's weights only
  in non-sensory rows; degree-preserving swaps exchange the targets of
  two existing edges, whose targets are never sensory, so no control
  gives a sensory group an input.
\item
  \textbf{Swaps.} N2 and N4 make 10·\textbar E\textbar{} swap attempts
  on their edge set; an attempt is rejected if the two edges share a
  source or a target or if either new edge already exists (for N4,
  existence is checked against the whole matrix, so interior swaps
  cannot duplicate a boundary edge). Accepted swaps are not counted by
  the engine; interior overlap with the connectome (0.111 to 0.121 for
  N4, 0.089 to 0.095 for N5) is the mixing diagnostic we report.
\item
  \textbf{N5.} For each non-sensory source, its interior outputs are
  moved to randomly chosen free rows among the non-sensory,
  non-descending groups; its boundary outputs stay in place.
\item
  \textbf{Kenyon-cell inputs.} The projection-neuron-to-Kenyon-cell
  matrix is randomised by degree-preserving swaps in N2, N4 and N5, and
  by shuffling which Kenyon cells receive each group's input in N1. The
  Kenyon-cell-to-group matrix and initial Kenyon-cell-to-output weights
  are shared.
\item
  \textbf{Boundary check.} For N4 and N5 the engine compares every
  boundary edge with the connectome before calibration and aborts if any
  differs (maximum absolute difference 0 in every construction).
\end{itemize}

\subsection{Calibration}\label{calibration}

All experiments here use \textbf{global} calibration: W row-normalised
to unit absolute input, then a global gain and a Kenyon-cell gain found
by bisection so that mean group activity is 0.15 and mean Kenyon-cell
activity 0.05 over a fixed stimulus set. Convergence is exact in every
condition. Because row normalisation divides each row by its own total
input, a boundary edge that is bit-identical before calibration can
carry a slightly different share of weight afterwards; all path
statistics in Section \ref{sec-shortcuts} are computed on the calibrated
matrices that the runs actually used.

Global calibration matches the mean activity but not its spread: across
the first grid the connectome had between-group activity SD 0.163, 46
silent groups of 512 and 2.0 \% saturated, against SD 0.080 to 0.086, 8
to 19 silent and under 0.3 \% saturated for N1 to N3; N4 and N5
(corrected grid, 40 runs each) have median SD 0.087 (runs 0.081 to
0.095) and median 10.5 and 13 silent groups (runs 4 to 17). Paper-v1
repeated its main endpoints under a distribution-matched calibration
that equalises mean, spread and silent fraction, with the same
conclusions (Appendix \ref{app-v1}).

\begin{itemize}
\tightlist
\item
  \textbf{global}: W row-normalised to unit absolute input; one global
  gain and one Kenyon-cell gain found by bisection so mean group
  activity = 0.15 and mean Kenyon-cell activity = 0.05 over a fixed
  stimulus ensemble (16 odors × 2 concentrations, single-channel
  stimuli, one mixed stimulus). Exact convergence for all conditions
  (relative error 0).
\item
  \textbf{pergroup}: per-group multiplicative gain feedback, 150
  iterations (the exploratory-phase scheme). Residuals differ by
  condition (connectome 0.23, nulls 0.02--0.07).
\item
  \textbf{rawglobal}: global gain without structural normalisation;
  circuit analysis only.
\item
  \textbf{distmatch}: structural normalisation plus per-group quantile
  matching to a fixed log-normal reference profile (mean 0.15, SD/mean
  0.8), equalising mean, SD and silent fraction. Achieved: mean
  0.148--0.149, SD 0.118, sorted-profile error 0.0013 in all four
  conditions.
\end{itemize}

\section{Evaluation and statistics}\label{app-eval}

Every 50 generations the 16 agents with the highest training fitness per
island are evaluated for 8 further lifetimes each in the run's own
ecology. The random-number generator is reset to seed 777 before the
batch, so conditions and ablations face the same sequence of generated
worlds and the comparison is paired at the level of the batch, not of
the individual arena. Evaluation worlds are drawn from the same
distribution as training worlds and are not held out. The run-level
number is the mean over the 128 lifetimes. Five versions run in
parallel: intact, plasticity blocked, olfaction blocked, vision blocked,
and all read-outs zeroed. Ablation cost is intact minus ablated. The
registered primary endpoint is the time average of this fitness over
generations 0 to 550, computed by the trapezoid rule and divided by the
generation span; the fitness at the last of these probes is a registered
secondary endpoint. Because the probes are 50 generations apart, that
last probe is at generation 550, while evolution itself continues to
generation 600; wherever we report an end-of-run difference it is
measured at generation 550. Drafts up to v3.2 labelled these numbers
``generation 600'', which was wrong; the values were always the last
probe and are unchanged.

Seeds are the unit of replication. Within an ecology we test paired
differences (connectome minus control) with two-sided Wilcoxon
signed-rank tests, Holm-corrected across controls, and report medians
with percentile bootstrap intervals over resampled seeds (10,000
resamples). Runs of the same seed share their random controls across
ecologies, so statements across ecologies average each seed's ecologies
first and test the 10 seed means; the estimand is then the median across
seeds of a seed's equally weighted mean difference. Counts over all 40
runs are descriptive.

For claims that a difference is absent we ask for equivalence rather
than a non-significant p-value: the 90 \% bootstrap interval of the
median paired difference must lie inside ±δ with δ = 0.10 fitness units.
The bound was chosen after the first grid, when the size of the shortcut
effect was known (the smallest shortcut effect we wanted to exclude was
0.20) and before any equivalence interval was computed; it is a claim
about what is small relative to that effect, not about what is
biologically negligible. Equivalence is claimed only where we state it,
and only for the estimand above; a non-significant test elsewhere is
reported as no detected difference. Analyses added during AI-assisted
review of earlier drafts (Appendix \ref{app-ai}) are labelled post hoc.

\section{Pre-registration record and the noise
correction}\label{app-prereg}

The first grid was run with the heading-noise term written as 0.05 ξ
with ξ \textasciitilde{} U(0, 1), which adds a steady 0.025 rad per
step; because the turn read-out subtracts a running mean of the
left--right difference, no network could offset the drift. The error
came to light through a question about that term during AI-assisted
review of the first draft. Re-evaluating the stored first-grid
populations with zero-mean noise showed that connectome populations lost
0.174 fitness when the drift was removed while controls lost 0.00 to
0.04 (10 of 10 seeds, p\_holm = 0.0059), and that the connectome's
generation-0 lead over N1 and N2 became non-significant; the late
overtaking by N1 and N2 was unchanged. Every experiment after the first
grid uses the zero-mean form given in Appendix \ref{app-model}. The
first grid is reported in Appendix \ref{app-first} because it motivated
the rest and because the overtaking it shows is the effect that the
later experiments explain; its generation-0 result is not used.

The protocols were frozen in this order: \texttt{PROTOCOL\_ECO.md}
before the first grid, \texttt{PROTOCOL\_MUT.md} after its analysis,
\texttt{PROTOCOL\_FIX.md} (with the boundary-preserving controls added
in amendment 1) before the corrected grid, and
\texttt{PROTOCOL\_SWAP.md} before the intervention runs and before the
corrected grid was analysed. The research log in the repository records
the exploratory phase that preceded the first protocol, including the
design changes made then; none of those affect the frozen experiments
reported here.

\subsection{Design changes and corrections, in
order}\label{design-changes-and-corrections-in-order}

The main text reports one correction made after pre-registration
(Appendix \ref{app-prereg}). For completeness, this section lists every
design change that reached the frozen protocols or the reported results,
in the order they were made; the full dated log is
\texttt{RESEARCH\_LOG.md} in the repository.

{
\begin{longtable}[]{@{}
  >{\raggedright\arraybackslash}p{(\linewidth - 4\tabcolsep) * \real{0.3333}}
  >{\raggedright\arraybackslash}p{(\linewidth - 4\tabcolsep) * \real{0.3333}}
  >{\raggedright\arraybackslash}p{(\linewidth - 4\tabcolsep) * \real{0.3333}}@{}}
\toprule\noalign{}
\begin{minipage}[b]{\linewidth}\raggedright
when
\end{minipage} & \begin{minipage}[b]{\linewidth}\raggedright
what
\end{minipage} & \begin{minipage}[b]{\linewidth}\raggedright
effect on reported results
\end{minipage} \\
\midrule\noalign{}
\endhead
\bottomrule\noalign{}
\endlastfoot
exploratory phase, before \texttt{PROTOCOL.md} & 27 changes to brain
construction, calibration and world code (for example: mandatory
inclusion of the gustatory-to-dopaminergic path groups; replacement of
clustered Kenyon cells by individually wired ones; removal of an energy
loophole that let agents survive without eating; a retracted
learning-specificity result traced to a numerically unstable
calibration) & none of the frozen experiments were run before these
changes \\
after \texttt{PROTOCOL\_ECO.md} was frozen, before any run & the setting
meant to remove the predator instead enabled an adaptive predator; fixed
and re-frozen before the first grid & none \\
after the first grid & stored genomes were population samples rather
than elites, by an indexing error & affects only the post hoc assays of
Appendix \ref{app-checks}, as stated there \\
after the first grid (AI-assisted review of draft 1) & one-sided heading
noise, Appendix \ref{app-prereg} & the first grid's generation-0 result
is not used; all later experiments corrected \\
after the first grid & the standard nulls carry olfactory-to-motor
shortcuts, Section \ref{sec-shortcuts} & motivated the corrected grid
and the intervention \\
\end{longtable}
}

\section{First grid}\label{app-first}

\begin{figure}
\centering
\includegraphics[width=1\linewidth,height=\textheight,keepaspectratio,alt={First grid: common-garden fitness over 600 generations in the four ecologies, median and interquartile range over 10 seeds. This experiment was run before the noise correction (Appendix ) and uses controls that carry sensory-to-motor shortcuts.}]{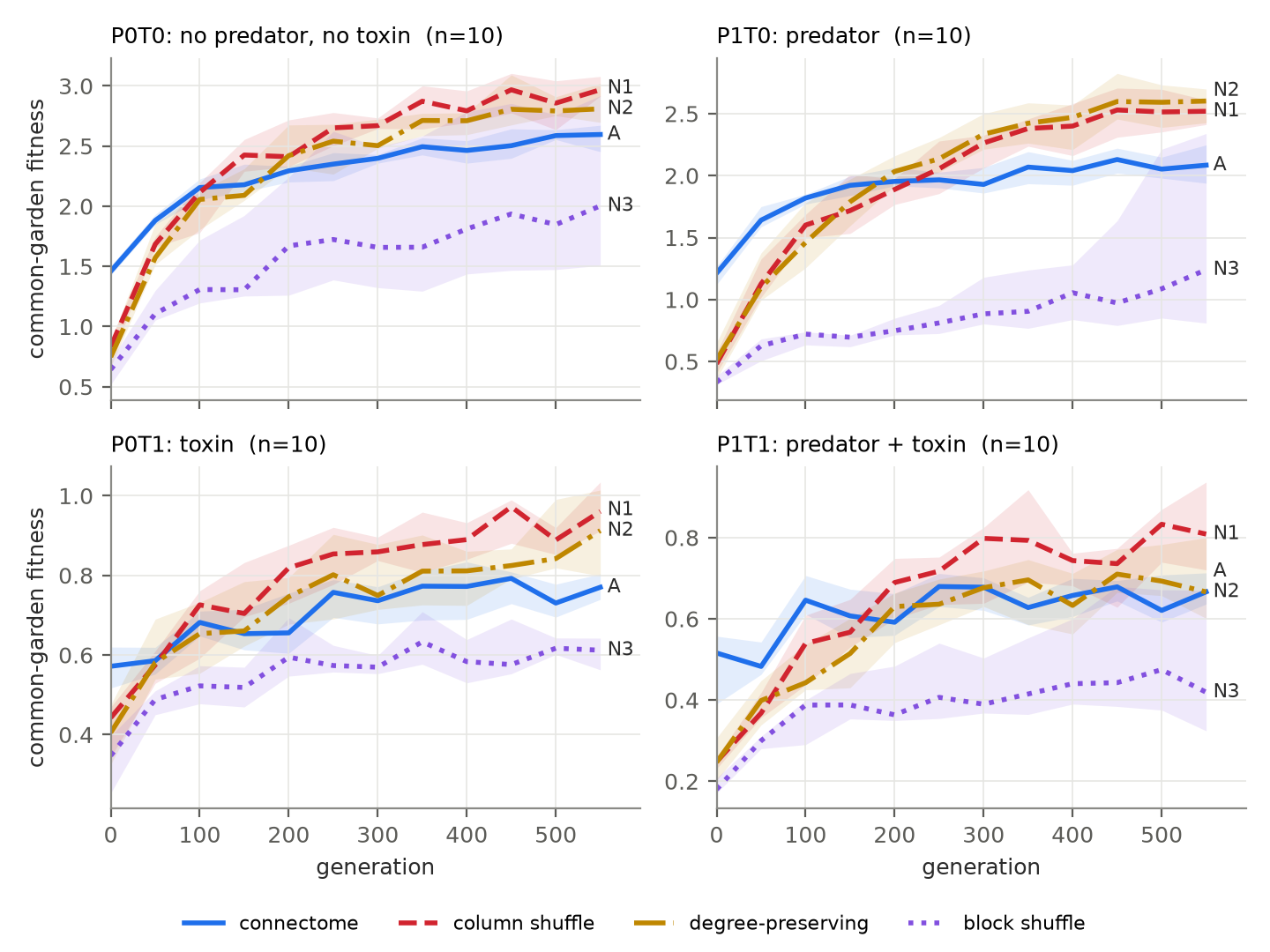}
\caption{First grid: common-garden fitness over 600 generations in the
four ecologies, median and interquartile range over 10 seeds. This
experiment was run before the noise correction (Appendix
\ref{app-prereg}) and uses controls that carry sensory-to-motor
shortcuts.}\label{fig:first}
\end{figure}

The first grid (pre-registered; 4 ecologies × 10 seeds × 600
generations; conditions connectome, N1, N2, N3; noise error present,
Appendix \ref{app-prereg}) reproduced the pattern that motivates this
paper. Connectome populations started ahead of all three controls in
every ecology (12 of 12 comparisons, p\_holm ≤ 0.027) and were then
overtaken. At the last probe the column shuffle N1 was ahead in every
ecology (p\_holm 0.006 to 0.041) and, on seed means across ecologies, in
all 10 seeds (p\_holm = 0.0059); the degree-preserving control N2 was
ahead in 8 of 10 seeds (p\_holm = 0.0195). The registered primary
endpoint, time-averaged fitness, showed no connectome advantage over
either (seed-mean medians −0.10 against N1 and −0.03 against N2).
Crossings were sustained in 36 of 40 runs for N1 and 31 of 40 for N2
(Figure \ref{fig:first}). Connectome populations also depended on
olfaction far less than the controls in all 12 ecology × control
comparisons (p\_holm = 0.0059 each).

One explanation can be set aside. In the mutational-neighbourhood assay
the fraction of one-step mutants beating their parent's clone
distribution was the same for connectome and control genomes (9.6 \%
against 9.4 \% for N1 and 9.8 \% for N2 at generation 0; 7.0 \% against
7.0 \% and 6.4 \% at generation 500), with run-level differences whose
95 \% intervals all lie inside ±2.7 percentage points; at the ±2
percentage-point margin set in that analysis, 12 of the 18 comparisons
fall inside and 6 do not, so this is similarity at a coarse resolution
rather than demonstrated equivalence. This is a statement about one-step
accessibility at this resolution, not about evolvability in general: it
does not exclude differences in the size of the best mutations, in
multi-step accessibility, or in epistasis. Connectome genomes did lose
more to large mutations (10th percentile −0.87 of parent fitness against
−0.77 to −0.81, p\_holm 0.007 to 0.012 at generation 0).

\section{Corrected grid: further results}\label{app-grid}

We re-ran the grid with the noise term corrected and with N4 and N5 in
place of N3 (\texttt{PROTOCOL\_FIX.md}; 40 runs, 5 conditions each, 600
generations). Table \ref{tbl:grid} gives differences at the last probe.

Against the shortcut-carrying controls the connectome is behind (p\_holm
= 0.029 for N1 and 0.016 for N2 on seed means). Against the
boundary-preserving controls the seed-mean difference lies inside the
equivalence bound for both, +0.002 for N4 and −0.074 for N5, the latter
close to the edge. Per ecology the estimates are not uniform and we do
not claim equivalence there: in the safe-food ecology N5 is ahead of the
connectome by 0.26 (p\_holm = 0.029) and in the predator-plus-toxin
ecology the connectome is ahead of N4 by 0.12, with an interval spanning
zero. Ten further seeds in the two T0 ecologies, run under rules fixed
beforehand (Appendix \ref{app-n20}), did not replicate that lead (seeds
10 to 19: +0.031, 90 \% CI −0.068 to +0.152; all 20 seeds: −0.104,
−0.208 to +0.016, p\_holm = 0.27). At n = 20 the per-ecology intervals
for N4 and N5 lie inside ±0.10 in P1T0 but not in P0T0; N1 stays ahead
in both (p\_holm ≤ 0.004), N2 stays ahead in P1T0 (p\_holm = 0.046) but
its P0T0 lead shrinks to −0.214 (−0.375 to +0.042, p\_holm = 0.25). The
registered primary endpoint gives seed-mean medians of −0.03 (N1), −0.02
(N2), +0.03 (N4) and −0.01 (N5), none significant, and we report these
as no detected difference rather than as equivalence.

Because δ was fixed after the first grid (Appendix \ref{app-eval}), we
also report, post hoc, the smallest symmetric bound each seed-mean
interval satisfies (Appendix \ref{app-delta}): 0.045 for N4 and 0.096
for N5. The N4 verdict therefore holds for any bound down to 0.05; the
N5 verdict holds at 0.10 but not at 0.075 and should be read as
marginal.

The sensory-strategy difference behaves the same way. The cost of
blocking olfaction is 0.03 for the connectome against 0.68 for N1 and
0.64 for N2 (p\_holm = 0.008), and 0.10 for N4 and 0.17 for N5. The
differences from the connectome are −0.07 {[}−0.11, −0.02{]} for N4 and
−0.11 {[}−0.22, +0.02{]} for N5, neither significant after Holm
correction across the four controls (p\_holm = 0.13 for both), and five-
to ninefold smaller than the 0.6 differences against N1 and N2. For
vision the corresponding differences are +0.11 {[}+0.03, +0.22{]}
against N1 and +0.20 {[}+0.11, +0.32{]} against N2, both significant,
and +0.03 {[}−0.25, +0.15{]} against N4 and −0.04 {[}−0.18, −0.00{]}
against N5. With the noise term corrected and the boundary matched, the
connectome no longer starts ahead either: the generation-0 seed-mean
difference against N4 is −0.18 (p\_holm = 0.15), so if anything N4
starts ahead.

\begin{figure}
\centering
\includegraphics[width=1\linewidth,height=\textheight,keepaspectratio,alt={Corrected grid: connectome, shortcut-carrying controls (N1, N2) and boundary-preserving controls (N4, N5).}]{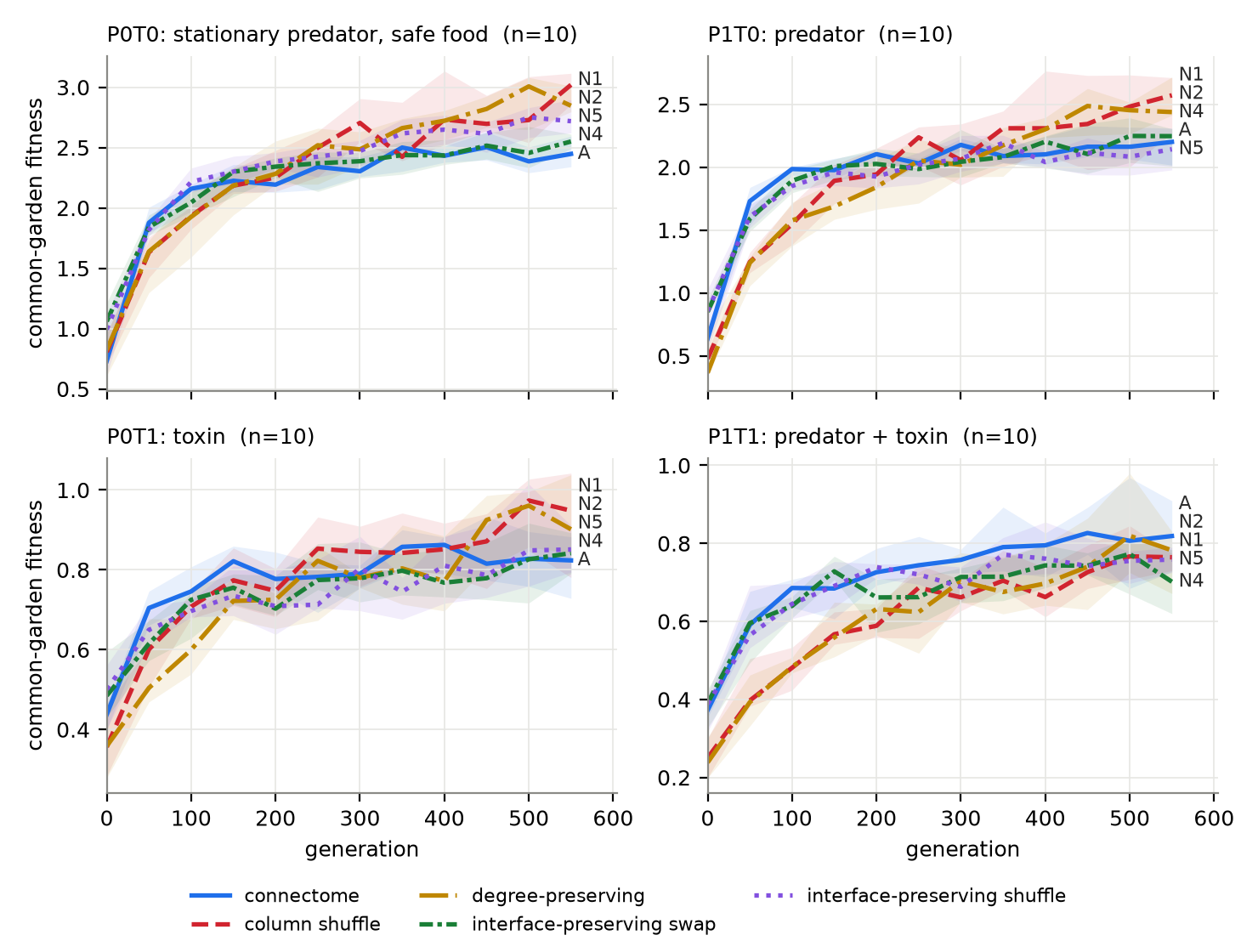}
\caption{Corrected grid: connectome, shortcut-carrying controls (N1, N2)
and boundary-preserving controls (N4, N5).}\label{fig:grid}
\end{figure}

\subsection{Sensitivity of the equivalence verdicts to the
bound}\label{app-delta}

The bound δ = 0.10 was fixed after the first grid (Appendix
\ref{app-eval}). Here each registered 90 \% interval of the corrected
grid (connectome minus control, last probe) is read against other
bounds. The smallest symmetric bound an interval satisfies is the larger
of its two ends in absolute value. The intervals are read from the file
Table \ref{tbl:grid} is taken from (\texttt{results/equivalence.txt}) by
\texttt{analyze\_delta\_sens.py}; nothing is recomputed. Equivalence is
claimed in the paper only on seed means; the per-ecology rows are shown
to make the heterogeneity visible, not as claims.

{
\begin{longtable}[]{@{}
  >{\raggedright\arraybackslash}p{(\linewidth - 18\tabcolsep) * \real{0.1370}}
  >{\raggedright\arraybackslash}p{(\linewidth - 18\tabcolsep) * \real{0.0959}}
  >{\raggedright\arraybackslash}p{(\linewidth - 18\tabcolsep) * \real{0.0959}}
  >{\raggedright\arraybackslash}p{(\linewidth - 18\tabcolsep) * \real{0.2192}}
  >{\raggedright\arraybackslash}p{(\linewidth - 18\tabcolsep) * \real{0.1096}}
  >{\raggedright\arraybackslash}p{(\linewidth - 18\tabcolsep) * \real{0.0685}}
  >{\raggedright\arraybackslash}p{(\linewidth - 18\tabcolsep) * \real{0.0685}}
  >{\raggedright\arraybackslash}p{(\linewidth - 18\tabcolsep) * \real{0.0685}}
  >{\raggedright\arraybackslash}p{(\linewidth - 18\tabcolsep) * \real{0.0685}}
  >{\raggedright\arraybackslash}p{(\linewidth - 18\tabcolsep) * \real{0.0685}}@{}}
\toprule\noalign{}
\begin{minipage}[b]{\linewidth}\raggedright
cell
\end{minipage} & \begin{minipage}[b]{\linewidth}\raggedright
control
\end{minipage} & \begin{minipage}[b]{\linewidth}\raggedright
median
\end{minipage} & \begin{minipage}[b]{\linewidth}\raggedright
90 \% CI
\end{minipage} & \begin{minipage}[b]{\linewidth}\raggedright
smallest bound
\end{minipage} & \begin{minipage}[b]{\linewidth}\raggedright
δ = 0.05
\end{minipage} & \begin{minipage}[b]{\linewidth}\raggedright
0.075
\end{minipage} & \begin{minipage}[b]{\linewidth}\raggedright
0.10
\end{minipage} & \begin{minipage}[b]{\linewidth}\raggedright
0.15
\end{minipage} & \begin{minipage}[b]{\linewidth}\raggedright
0.20
\end{minipage} \\
\midrule\noalign{}
\endhead
\bottomrule\noalign{}
\endlastfoot
seed mean & N4 & +0.002 & {[}−0.026, +0.045{]} & 0.045 & yes & yes & yes
& yes & yes \\
seed mean & N5 & −0.074 & {[}−0.096, +0.002{]} & 0.096 & no & no & yes &
yes & yes \\
P0T0 & N4 & −0.064 & {[}−0.255, +0.002{]} & 0.255 & no & no & no & no &
no \\
P0T0 & N5 & −0.264 & {[}−0.368, −0.179{]} & 0.368 & no & no & no & no &
no \\
P1T0 & N4 & +0.021 & {[}−0.026, +0.038{]} & 0.038 & yes & yes & yes &
yes & yes \\
P1T0 & N5 & +0.067 & {[}−0.067, +0.090{]} & 0.090 & no & no & yes & yes
& yes \\
P0T1 & N4 & −0.031 & {[}−0.116, +0.085{]} & 0.116 & no & no & no & yes &
yes \\
P0T1 & N5 & −0.015 & {[}−0.102, +0.061{]} & 0.102 & no & no & no & yes &
yes \\
P1T1 & N4 & +0.120 & {[}−0.064, +0.283{]} & 0.283 & no & no & no & no &
no \\
P1T1 & N5 & +0.070 & {[}−0.009, +0.205{]} & 0.205 & no & no & no & no &
no \\
\end{longtable}
}

On seed means the N4 verdict holds for every bound down to 0.05. The N5
verdict holds at the registered 0.10 but not at 0.075: it is marginal,
and the paper says so. Per ecology the smallest bound ranges from 0.038
(N4, P1T0) to 0.368 (N5, P0T0).

\subsection{Distribution-matched calibration}\label{app-dist}

Frozen before the runs: protocol \texttt{PROTOCOL\_DIST2.md}
(\texttt{0124b50ad5345279}), analysis \texttt{analyze\_dist2.py}
(\texttt{d12b803c141faa1e}, unchanged at analysis), the corrected grid's
frozen engine \texttt{evo\_fix.py} and flags with
\texttt{-\/-homeo\ distmatch} in place of \texttt{global}. All 40 runs
(4 ecologies × seeds 0--9) finished; each configuration equals its
global-calibration counterpart apart from run name and calibration.
Output \texttt{results/dist2\_key.json}.

\textbf{Manipulation check} (registered: at least 36 of 40 runs with
\textbar SD − SD\_A\textbar{} ≤ 0.01, \textbar silent −
silent\_A\textbar{} ≤ 10 and \textbar saturated − saturated\_A\textbar{}
≤ 0.02 for N4 and N5): 40/40 runs → PASSED. Calibrated activity by
condition, medians over the 40 runs:

{
\begin{longtable}[]{@{}llll@{}}
\toprule\noalign{}
condition & between-group SD & silent groups & Kenyon-cell mean \\
\midrule\noalign{}
\endhead
\bottomrule\noalign{}
\endlastfoot
A & 0.118 & 7 & 0.050 \\
N1 & 0.118 & 3 & 0.050 \\
N2 & 0.118 & 6.5 & 0.050 \\
N4 & 0.118 & 4.5 & 0.050 \\
N5 & 0.118 & 4.5 & 0.050 \\
\end{longtable}
}

Under global calibration the same quantities were SD 0.163 / 46 silent
groups for the connectome against 0.080--0.087 / 8--19 for the controls
(main text); distribution matching removes that difference. N1's
Kenyon-cell mean ranged 0.038--0.050 (a known limitation of the scheme
for the column shuffle).

\textbf{P1 and P2 --- seed means over four ecologies, median with 90 \%
bootstrap interval:}

{
\begin{longtable}[]{@{}
  >{\raggedright\arraybackslash}p{(\linewidth - 6\tabcolsep) * \real{0.2500}}
  >{\raggedright\arraybackslash}p{(\linewidth - 6\tabcolsep) * \real{0.2500}}
  >{\raggedright\arraybackslash}p{(\linewidth - 6\tabcolsep) * \real{0.2500}}
  >{\raggedright\arraybackslash}p{(\linewidth - 6\tabcolsep) * \real{0.2500}}@{}}
\toprule\noalign{}
\begin{minipage}[b]{\linewidth}\raggedright
contrast
\end{minipage} & \begin{minipage}[b]{\linewidth}\raggedright
global (registered grid)
\end{minipage} & \begin{minipage}[b]{\linewidth}\raggedright
distribution-matched
\end{minipage} & \begin{minipage}[b]{\linewidth}\raggedright
verdict
\end{minipage} \\
\midrule\noalign{}
\endhead
\bottomrule\noalign{}
\endlastfoot
A − N4 & +0.002 {[}−0.026, +0.045{]} & +0.055 {[}+0.035, +0.097{]} &
EQUIVALENT \\
A − N5 & −0.074 {[}−0.096, +0.002{]} & −0.009 {[}−0.076, +0.046{]} &
EQUIVALENT \\
A − N1 & −0.224 {[}−0.337, −0.103{]} & −0.274 {[}−0.306, −0.123{]} &
BEHIND \\
A − N2 & −0.199 {[}−0.321, −0.133{]} & −0.146 {[}−0.231, −0.037{]} &
BEHIND \\
\end{longtable}
}

\textbf{S1 --- change caused by calibration, seed mean of (A − N) under
distribution matching minus (A − N) under global:}

{
\begin{longtable}[]{@{}lll@{}}
\toprule\noalign{}
control & change & verdict \\
\midrule\noalign{}
\endhead
\bottomrule\noalign{}
\endlastfoot
N1 & +0.010 {[}−0.112, +0.072{]} & UNDETERMINED \\
N2 & +0.101 {[}+0.016, +0.133{]} & CHANGED \\
N4 & +0.062 {[}−0.002, +0.117{]} & UNDETERMINED \\
N5 & +0.076 {[}−0.017, +0.128{]} & UNDETERMINED \\
\end{longtable}
}

\textbf{S2 --- N5 ahead of the connectome in P0T0 under distribution
matching:} A − N5 = −0.066 {[}−0.320, +0.124{]} → NOT REPLICATED.

\textbf{Per ecology (descriptive), A minus control, median {[}90 \%
CI{]}:}

{
\begin{longtable}[]{@{}llll@{}}
\toprule\noalign{}
ecology & control & global & distribution-matched \\
\midrule\noalign{}
\endhead
\bottomrule\noalign{}
\endlastfoot
P0T0 & N1 & −0.621 {[}−0.661, −0.469{]} & −0.437 {[}−0.563, −0.235{]} \\
P0T0 & N2 & −0.403 {[}−0.421, −0.308{]} & −0.478 {[}−0.677, −0.052{]} \\
P0T0 & N4 & −0.064 {[}−0.255, +0.002{]} & +0.139 {[}−0.128, +0.230{]} \\
P0T0 & N5 & −0.264 {[}−0.368, −0.179{]} & −0.066 {[}−0.320, +0.124{]} \\
P1T0 & N1 & −0.312 {[}−0.441, −0.197{]} & −0.489 {[}−0.664, −0.275{]} \\
P1T0 & N2 & −0.338 {[}−0.530, −0.074{]} & −0.164 {[}−0.483, −0.013{]} \\
P1T0 & N4 & +0.021 {[}−0.026, +0.038{]} & +0.051 {[}−0.120, +0.144{]} \\
P1T0 & N5 & +0.067 {[}−0.067, +0.090{]} & +0.041 {[}−0.204, +0.214{]} \\
P0T1 & N1 & −0.159 {[}−0.205, +0.010{]} & −0.096 {[}−0.182, +0.010{]} \\
P0T1 & N2 & −0.017 {[}−0.333, +0.005{]} & −0.020 {[}−0.122, +0.096{]} \\
P0T1 & N4 & −0.031 {[}−0.116, +0.085{]} & +0.073 {[}−0.066, +0.104{]} \\
P0T1 & N5 & −0.015 {[}−0.102, +0.061{]} & +0.036 {[}−0.086, +0.089{]} \\
P1T1 & N1 & +0.054 {[}−0.054, +0.127{]} & +0.063 {[}−0.040, +0.099{]} \\
P1T1 & N2 & +0.015 {[}−0.132, +0.137{]} & +0.104 {[}−0.091, +0.111{]} \\
P1T1 & N4 & +0.120 {[}−0.064, +0.283{]} & −0.012 {[}−0.099, +0.073{]} \\
P1T1 & N5 & +0.070 {[}−0.009, +0.210{]} & −0.027 {[}−0.066, +0.078{]} \\
\end{longtable}
}

\subsection{Extension to 20 seeds in two ecologies}\label{app-n20}

Seeds 10 to 19 of the two T0 ecologies were run with the corrected
grid's frozen engine and flags after rules R1 to R3 were written
(research log 46; \texttt{analyze\_n20.py}). Every one of the 40 runs
has the same configuration as seed 0 apart from name and seed. Endpoint:
fitness at the last common-garden probe (generation 550); connectome
minus control, median with 90 \% bootstrap interval; p\_holm from
Wilcoxon signed-rank tests at n = 20, Holm-corrected across the four
controls within each ecology.

{
\begin{longtable}[]{@{}
  >{\raggedright\arraybackslash}p{(\linewidth - 12\tabcolsep) * \real{0.1429}}
  >{\raggedright\arraybackslash}p{(\linewidth - 12\tabcolsep) * \real{0.1429}}
  >{\raggedright\arraybackslash}p{(\linewidth - 12\tabcolsep) * \real{0.1429}}
  >{\raggedright\arraybackslash}p{(\linewidth - 12\tabcolsep) * \real{0.1429}}
  >{\raggedright\arraybackslash}p{(\linewidth - 12\tabcolsep) * \real{0.1429}}
  >{\raggedright\arraybackslash}p{(\linewidth - 12\tabcolsep) * \real{0.1429}}
  >{\raggedright\arraybackslash}p{(\linewidth - 12\tabcolsep) * \real{0.1429}}@{}}
\toprule\noalign{}
\begin{minipage}[b]{\linewidth}\raggedright
ecology
\end{minipage} & \begin{minipage}[b]{\linewidth}\raggedright
control
\end{minipage} & \begin{minipage}[b]{\linewidth}\raggedright
seeds 0--9
\end{minipage} & \begin{minipage}[b]{\linewidth}\raggedright
seeds 10--19
\end{minipage} & \begin{minipage}[b]{\linewidth}\raggedright
n = 20
\end{minipage} & \begin{minipage}[b]{\linewidth}\raggedright
p\_holm (n = 20)
\end{minipage} & \begin{minipage}[b]{\linewidth}\raggedright
inside ±0.10 (R1)
\end{minipage} \\
\midrule\noalign{}
\endhead
\bottomrule\noalign{}
\endlastfoot
P0T0 & N1 & −0.621 {[}−0.661, −0.469{]} & −0.304 {[}−0.513, −0.161{]} &
−0.491 {[}−0.624, −0.298{]} & 0.001 & no \\
P0T0 & N2 & −0.403 {[}−0.421, −0.308{]} & +0.100 {[}−0.068, +0.332{]} &
−0.214 {[}−0.375, +0.042{]} & 0.248 & no \\
P0T0 & N4 & −0.064 {[}−0.255, +0.002{]} & +0.090 {[}−0.109, +0.217{]} &
−0.011 {[}−0.114, +0.080{]} & 0.841 & no \\
P0T0 & N5 & −0.264 {[}−0.368, −0.179{]} & +0.031 {[}−0.068, +0.152{]} &
−0.104 {[}−0.208, +0.016{]} & 0.265 & no \\
P1T0 & N1 & −0.312 {[}−0.441, −0.197{]} & −0.417 {[}−0.522, +0.067{]} &
−0.333 {[}−0.482, −0.197{]} & 0.004 & no \\
P1T0 & N2 & −0.338 {[}−0.530, −0.074{]} & −0.101 {[}−0.335, +0.103{]} &
−0.203 {[}−0.430, −0.032{]} & 0.046 & no \\
P1T0 & N4 & +0.021 {[}−0.026, +0.038{]} & −0.027 {[}−0.168, +0.050{]} &
+0.003 {[}−0.042, +0.029{]} & 1.000 & yes \\
P1T0 & N5 & +0.067 {[}−0.067, +0.090{]} & +0.003 {[}−0.175, +0.144{]} &
+0.056 {[}−0.097, +0.090{]} & 1.000 & yes \\
\end{longtable}
}

\textbf{R2 (replication of N5's lead in P0T0 in seeds 10 to 19):} A − N5
= +0.031 {[}−0.068, +0.152{]} → \textbf{NOT REPLICATED}.

\textbf{R3 (two-ecology pool, seed means over P0T0 and P1T0, n = 20; a
different estimand from Table \ref{tbl:grid}'s four-ecology seed
means):}

{
\begin{longtable}[]{@{}
  >{\raggedright\arraybackslash}p{(\linewidth - 6\tabcolsep) * \real{0.2500}}
  >{\raggedright\arraybackslash}p{(\linewidth - 6\tabcolsep) * \real{0.2500}}
  >{\raggedright\arraybackslash}p{(\linewidth - 6\tabcolsep) * \real{0.2500}}
  >{\raggedright\arraybackslash}p{(\linewidth - 6\tabcolsep) * \real{0.2500}}@{}}
\toprule\noalign{}
\begin{minipage}[b]{\linewidth}\raggedright
control
\end{minipage} & \begin{minipage}[b]{\linewidth}\raggedright
median {[}90 \% CI{]}
\end{minipage} & \begin{minipage}[b]{\linewidth}\raggedright
seeds with connectome ahead
\end{minipage} & \begin{minipage}[b]{\linewidth}\raggedright
inside ±0.10
\end{minipage} \\
\midrule\noalign{}
\endhead
\bottomrule\noalign{}
\endlastfoot
N1 & −0.401 {[}−0.497, −0.241{]} & 3/20 & no \\
N2 & −0.155 {[}−0.273, −0.060{]} & 4/20 & no \\
N4 & 0.000 {[}−0.087, +0.083{]} & 10/20 & yes \\
N5 & −0.059 {[}−0.093, +0.018{]} & 7/20 & yes \\
\end{longtable}
}

\textbf{Post hoc checks on the extension (not registered;
\texttt{analyze\_posthoc\_n20.py}).} The nulls of seeds 10 to 19 were
built like those of seeds 0 to 9: the olfactory-to-descending direct
share of the calibrated weights the runs used was 10.41 \% (median) for
N2 against 10.63 \% in seeds 0 to 9, and 0.01 \% for N4 and N5. Across
the 20 null realisations per ecology this share ranged from about 5 to
17 \% but did not predict the null's lead over the connectome (Spearman
\textbar ρ\textbar{} ≤ 0.21, p ≥ 0.38 for N1 and N2 in each ecology;
pooled ρ = −0.00 and −0.09). The interventions changed the connectome's
share from 0.01 \% to 1 to 10 \%, a range these realisations do not
cover. A rough plug-in simulation from the observed differences suggests
that per-ecology equivalence in P0T0 would need on the order of 100
seeds for N4 and is unlikely for N5 at any sample size, so we did not
extend further.

\subsection{Additional analyses (post
hoc)}\label{additional-analyses-post-hoc}

Full outputs: \texttt{results/eco\_analysis.txt} (registered tests),
\texttt{results/eco\_cross.txt} (crossing, cross-ecology, H3),
\texttt{results/eco\_pooled.txt} (across-ecology summaries),
\texttt{results/mut\_analysis.txt} (mutational neighbourhood with
confidence intervals), \texttt{results/revision\_analysis.txt}
(sustained crossing, fitness components, absolute ablation costs,
population samples against top 16), \texttt{results/eco\_calib.txt}
(activity diagnostics per run).

\section{Intervention and dose-response: further
details}\label{app-dose}

\subsection{Intervention}\label{intervention}

The pre-registered intervention (\texttt{PROTOCOL\_SWAP.md}; 20 runs in
the two ecologies with the clearest overtaking, corrected noise) rewires
the shortcut in both directions while preserving degrees and Dale signs.

Adding shortcuts to the connectome raised evolved fitness in every seed:
AS minus connectome at the last probe was +0.44 on seed means (10 of 10
seeds, p = 0.002; +0.445 in P0T0 and +0.475 in P1T0), and time-averaged
fitness +0.26 (10 of 10, p = 0.002). The effect grows over evolution: at
generation 0 the difference is +0.04 (p = 0.32, 90 \% CI −0.04 to +0.18,
wider than the equivalence bound, so an initial effect is not excluded),
and the increase from generation 0 to the last probe is +0.40 (10 of 10
seeds, p = 0.002). The intervention also transferred the sensory
strategy: the cost of blocking olfaction rose from 0.15 to 0.99 (p =
0.002), close to the shortcut-carrying control's 0.84.

Removing the shortcut from the degree-preserving control cost it 0.43
fitness in the safe-food ecology (p = 0.004) and an estimated 0.00 in
the predator ecology (the difference between ecologies is −0.36, p =
0.027). In neither ecology could we detect a difference between the
shortcut-free control and the connectome (P0T0 −0.11, 90 \% CI −0.23 to
+0.25; P1T0 +0.05, 90 \% CI −0.24 to +0.23), and its olfactory
dependence fell from 0.84 to 0.58 (p = 0.027). In this experiment the
degree-preserving control's own advantage over the connectome was +0.55
in P0T0 (p = 0.006) and +0.10 in P1T0 (p = 0.23), smaller in the
predator ecology than the corrected grid's +0.34: controls are drawn
from a single random stream in the order the conditions are listed, so
the intervention runs use different N2 realisations than the corrected
grid, and this is the size of the variation between realisations.

Shortcut-creating rewiring is therefore sufficient to produce the
control's advantage and to transfer its olfactory strategy, and
shortcut-removing rewiring removes the advantage where it was clear.
Because both interventions also rewire edges that are not shortcuts,
path shortening is the proposed mechanism rather than an isolated
variable. Section \ref{sec-dose} reports an interior-only rewiring
control, which does not come out clean, and Section \ref{sec-bsham} a
boundary-targeted one, which separates the shortcut edges from other
rewiring at the boundary.

\subsection{Dose-response and sham}\label{dose-response-and-sham}

The intervention in Section \ref{sec-swap} moved the shortcut share
between its two extremes and changed every rewired edge at once, so it
left two questions open: whether the effect follows the amount of
shortcut, and whether it needs the shortcut at all. We pre-registered
both (\texttt{PROTOCOL\_DOSE.md}; 2 ecologies × 10 seeds × 600
generations) and ran four graded doses, raising the connectome's
olfactory-to-descending share to at least 1, 3, 5 and 10 \% by
degree-preserving swaps, against a sham that performs 101 swaps on edge
pairs whose source is not an olfactory receptor and whose target is not
a descending group, leaving the shortcut share unchanged at 0.0001. Each
seed draws its own swaps, so the share reached and the swaps it took
vary across the ten seeds: 1.0 to 1.7 \% (3 to 27 swaps), 3.0 to 4.6 \%
(11 to 48), 5.0 to 5.8 \% (25 to 58) and 10.0 to 10.6 \% (68 to 129),
with no overlap in share between doses. The sham's 101 is the full
transplant's swap count on a fixed stream and is the same in every seed.

The comparison that separates the two variables is the full dose against
the sham: they rewire comparable amounts (the full dose took 68 to 129
swaps depending on the seed, mean 98.5, against the sham's 101) and
differ by four orders of magnitude in shortcut share. The full dose is
ahead of the sham by 0.534 on seed means (90 \% CI +0.373 to +0.693, 10
of 10 seeds; +0.525 in the safe-food ecology with 10 of 10 seeds and
+0.543 in the predator ecology with 9 of 10). The swap counts are not
matched seed by seed, and that strengthens the comparison rather than
weakening it: in the seven seeds where the full dose used fewer swaps
than the sham it was ahead in all seven, by 0.612 on average, against
0.352 in the three where it used more, and across seeds the gap does not
grow with the difference in swap count (r = −0.16). The number of
degree-preserving swaps therefore does not account for the effect; what
those swaps land on does.

Fitness also rose across the graded doses. On seed means over both
ecologies it went 2.37, 2.54, 2.69, 2.75, 2.93 from the connectome to
the 10 \% dose (Page's L, z = 4.68, p = 1.4 × 10\textsuperscript{−6};
per-seed Spearman correlation positive in 9 of 10 seeds, mean +0.74),
and the trend held separately in both ecologies (p = 1.5 ×
10\textsuperscript{−5} and 0.004). Dependence on olfaction rose with it:
the fitness lost when olfaction is blocked went from 0.28 to 1.26 in the
safe-food ecology and from 0.08 to 0.95 in the predator ecology (p = 3.4
× 10\textsuperscript{−5} and 5.6 × 10\textsuperscript{−7}). At
generation 0 the doses are indistinguishable (p = 0.98), so the graded
effect is a product of evolution rather than of the initial wiring. We
ran a sham only at the full swap count, so shortcut share and swap count
rise together across the intermediate doses and the shape of the curve
between them cannot separate the two; only its endpoint has a sham. The
boundary-targeted experiment (Section \ref{sec-bsham}, Appendix
\ref{app-bsham}) paired a sham with every dose, and the gap between dose
and sham grew with dose.

The sham itself did not behave as we predicted. We registered that sham
minus connectome would fall inside the ±0.10 equivalence bound, which
would have shown that rewiring by itself does nothing. It does not: on
seed means the difference is +0.034 with a 90 \% interval of −0.062 to
+0.123, which crosses the bound. The two ecologies also disagree in
sign, the sham being ahead by 0.144 in the safe-food ecology (90 \% CI
+0.072 to +0.222) and behind by 0.077 in the predator ecology (90 \% CI
−0.257 to +0.087), so this is an unstable effect rather than a small
consistent one. Following the interpretation registered for this
outcome, we report that some part of the advantage follows from
degree-preserving rewiring in general. It is a small part next to the
0.534 that the shortcut adds at the same swap count, and we cannot say
from these runs what it consists of.

The registered equivalence prediction was not supported. An unregistered
analysis of final-generation in-run fitness does meet the equivalence
criterion (Appendix \ref{app-inrun}), and does not change that
conclusion.

Taken with Section \ref{sec-swap}: at a fixed amount of
degree-preserving rewiring, whether that rewiring creates a
sensory-to-motor shortcut decides most of the difference, and it brings
the olfactory strategy with it. Rewiring that creates no shortcut is not
inert, but what it does is smaller and does not hold its sign across
ecologies.

\subsection{The sham equivalence test on an unregistered
outcome}\label{app-inrun}

\texttt{PROTOCOL\_DOSE.md} registered the common-garden probe of
Appendix \ref{app-eval} as the outcome, as \texttt{PROTOCOL\_ECO.md}
does for every experiment in this paper. On that outcome the registered
equivalence prediction for the sham fails: sham minus connectome is
+0.034 on seed means, 90 \% CI −0.062 to +0.123, crossing the ±0.10
bound (Section \ref{sec-dose}).

The engines also log an in-run fitness at each generation, evaluated in
the training worlds rather than in the common-garden batch, and the last
of those is generation 599. That outcome was not registered. On it the
same difference is −0.022, 90 \% CI −0.090 to +0.047, which lies inside
the bound and would have supported the prediction. We report this
because it is recoverable from the released logs and because the
direction of the disagreement is the one that flatters us; it does not
govern the registered test, which is the probe.

The disagreement is confined to the sham. Under the unregistered outcome
the graded series still rises monotonically (Page's L, z = 5.12, p = 1.5
× 10\textsuperscript{−7}, per-seed Spearman positive in 10 of 10 seeds)
and the generation-0 null still holds (p = 0.98). The swap-matched
comparison of Section \ref{sec-dose} is likewise unaffected in sign or
approximate size.

We did not investigate why the two outcomes differ for the sham
specifically. The probe resets its random stream and evaluates 16 agents
per island for 8 lifetimes in a fixed sequence of worlds, while the
in-run number is the population's fitness in the worlds it was selected
in; a condition whose advantage is partly specific to the worlds it
evolved in would show the two diverging, but we have not tested that
here.

\subsection{Boundary-targeted sham: full results and execution
record}\label{app-bsham}

Frozen before the runs (research log 54): protocol
\texttt{PROTOCOL\_BSHAM.md}, engine \texttt{evo\_bsham.py}
(\texttt{007fa61ff2c6f5c5}), analysis \texttt{analyze\_bsham.py}
(\texttt{cbb7f9e1c7aa6192}, unchanged at analysis). Output
\texttt{results/bsham\_key.json}. All 20 runs finished; the analysis's
pre-checks passed (configurations equal apart from run, seed and
predator speed; 80 boundary-sham constructions complete, shortcut share
unchanged, degrees kept;
\textbar w(Z→DN)\textbar/\textbar w(ORN→X)\textbar{} medians 0.92 to
1.05).

{
\begin{longtable}[]{@{}
  >{\raggedright\arraybackslash}p{(\linewidth - 6\tabcolsep) * \real{0.2500}}
  >{\raggedright\arraybackslash}p{(\linewidth - 6\tabcolsep) * \real{0.2500}}
  >{\raggedright\arraybackslash}p{(\linewidth - 6\tabcolsep) * \real{0.2500}}
  >{\raggedright\arraybackslash}p{(\linewidth - 6\tabcolsep) * \real{0.2500}}@{}}
\toprule\noalign{}
\begin{minipage}[b]{\linewidth}\raggedright
quantity (seed mean over P0T0 and P1T0, n = 10)
\end{minipage} & \begin{minipage}[b]{\linewidth}\raggedright
mean
\end{minipage} & \begin{minipage}[b]{\linewidth}\raggedright
90 \% CI
\end{minipage} & \begin{minipage}[b]{\linewidth}\raggedright
registered rule
\end{minipage} \\
\midrule\noalign{}
\endhead
\bottomrule\noalign{}
\endlastfoot
P: AS10 − BS10, last probe & +0.600 & {[}+0.414, +0.797{]} & lower bound
\textgreater{} +0.10 → shortcut-specific \\
R: AS10 − BS10, fresh worlds (4 × 8 lives, generation-599 elites) &
+0.639 & {[}+0.490, +0.792{]} & same category as P → robust \\
S2: AS10 − BS10, P0T0 / P1T0 & +0.568 / +0.631 & {[}+0.315, +0.832{]} /
{[}+0.428, +0.831{]} & descriptive \\
S3: BS10 − A & +0.007 & {[}−0.070, +0.086{]} & ±0.10 reference:
inside \\
S4: olfaction-ablation cost AS10 / BS10 & 1.175 / 0.354 & difference
+0.821 {[}+0.563, +1.079{]} & descriptive \\
S5: AS10 − A & +0.606 & {[}+0.435, +0.798{]} & lower bound
\textgreater{} 0 → replicated \\
\end{longtable}
}

S1 (AS\emph{k} − BS\emph{k} increases with dose): 0.158, 0.277, 0.357,
0.600 for 1, 3, 5, 10 \%; Page's L = 279, z = 3.18, p = 7.4 ×
10\textsuperscript{−4} → supported.

\textbf{Execution record.} Home workstation (M2 Max, 32 GB), one run at
a time. Three runs failed with GPU out-of-memory
(\texttt{bsham\_P0.2\_T0.0\_s1} twice, \texttt{bsham\_P0.0\_T0.0\_s2},
\texttt{bsham\_P0.0\_T0.0\_s5}); the causes were the reproduction step
holding several population-sized weight copies (peak 30.7 to 33 GB) and,
once, another service loading a 12 GB model on the same GPU. One
completed attempt was deleted by a driver that misread the log's last
line, and one attempt was stopped deliberately to test the memory fix.
Every affected run was re-run from scratch with identical flags; no
outcome was inspected before re-running (research logs 55 to 57 and 62).
The first four runs (\texttt{s0} both ecologies, \texttt{s1} P0T0,
\texttt{s2} P1T0) used \texttt{evo\_bsham.py}; the other 16 used
\texttt{evo\_bsham\_mem.py} (\texttt{f142f287e671b147}), which adds six
forced evaluations in reproduction and is otherwise identical; with
probes off it was bit-identical to \texttt{evo\_bsham.py} on the CPU
(peak memory 23.8 GB instead of 30.7 to 33 GB). As for every engine in
this project, re-running a seed reproduces the constructions and
generation 0 exactly and later generations only in distribution (see
Data and code availability).

\section{Further checks}\label{app-checks}

\subsection{Assays}\label{assays}

\textbf{Mutational neighbourhood.} For each run, checkpoint (generations
0, 250, 500) and stored genome, 32 unmutated clones and 32 one-step
mutants at 0.5×, 1× and 2× the mutation magnitudes, each evaluated 4
times in the home ecology. The clones give each parent's noise
distribution; the reported statistic is the fraction of mutants
exceeding the 95th percentile of their parent's clones, which is 5 \%
when mutations have no effect, together with the tails of the
mutant-minus-clone difference.

\textbf{Turning-noise control.} Generation-0 and generation-500
population samples of all 40 first-grid runs re-evaluated in their home
ecology with the original and with zero-mean heading noise.

\textbf{Interior necessity.} Generation-500 population samples of the
corrected grid re-evaluated intact, with interior weights zeroed, and
with interior weights randomly permuted within the interior mask. Losses
are expressed as (intact − perturbed)/intact per run; the smallest
intact fitness over all runs and conditions was 0.44, so the ratio is
well behaved.

\subsection{Interior perturbation}\label{interior-perturbation}

Re-evaluating generation-500 populations of the corrected grid with the
interior disturbed shows that they depend on it: zeroing interior
weights costs 18 to 23 \% of fitness and randomly permuting them costs
46 to 55 \%, in every condition, with no detected difference between the
connectome and any control (all p ≥ 0.16). This rules out the
possibility that the evolved agents run on the sensory-to-motor boundary
alone, which would have made the comparison in Section
\ref{sec-equivalence} uninformative. It does not show that the task
could not be solved by a simpler reflex; it shows that these agents did
not solve it that way.

\subsection{What survives from the first
experiment}\label{what-survives-from-the-first-experiment}

Of the first grid's findings, the connectome's generation-0 advantage
does not survive: it rested on the biased turning noise together with
the boundary difference, and is no longer detected once both are
addressed. The overtaking by N1 and N2 survives, and Section
\ref{sec-swap} accounts for it.

Two observations hold in both grids and are reported with their metrics
in Appendix \ref{app-v1}. We detected no odor-specific associative
learning in any condition: in an odor-reversal assay crossed with
plasticity blocking, the interaction never reached significance,
although plasticity itself was worth 0.09 to 0.30 fitness units. With 10
seeds a small interaction cannot be excluded. Randomised networks also
stayed structurally distinct from the connectome throughout: after 250
generations their support overlap with it had moved by at most 0.003,
the ipsilateral share of weight was 0.77 in the connectome against 0.50
in N1 and N2, and reciprocal edges 0.43 against 0.05 to 0.12. In
paper-v1 only (\texttt{results/evolution\_analysis.txt}, three
calibration schemes; not measured in the two grids), weights moved over
the same period by 0.5 to 3 \% of their initial spread while gains,
leaks and read-outs moved by about 6 to 11 initial standard deviations.
We report that as the measured pattern of change; attributing the
adaptation to those parameters would need a freezing experiment we did
not run.

\section{Earlier model version (paper-v1)}\label{app-v1}

The results in this appendix come from paper-v1, the first
pre-registered experiment (\texttt{PROTOCOL.md}: 4 conditions × 10 seeds
× 300 generations in one ecology with toxic probability 0.5, toxic-bite
loss 0.4 and a predator pursuing at 0.1·v\_max; common garden every 25
generations), run before the ecology grids and before the noise
correction. Stored genomes there are also population samples rather than
elites. It is included for completeness; the main text does not rest on
it.

Circuit properties of initial brains, 4 conditions × 10 seeds per
scheme:

\begin{itemize}
\tightlist
\item
  Kenyon-cell odor decorrelation favours the connectome under
  \textbf{pergroup} (10/10 seeds, p\_holm = 0.006) and favours the nulls
  under \textbf{global} and \textbf{rawglobal} (0--3/10 seeds).
\item
  Dopaminergic responses to sugar and malaise favour the connectome
  under \textbf{global} (10/10, p\_holm = 0.006), show no consistent
  difference under \textbf{pergroup}, and favour the nulls under
  \textbf{rawglobal}.
\item
  Mushroom-body learning specificity shows no significant difference in
  any scheme.
\item
  Looming-escape turn sign is correct in the connectome in 10/10 seeds
  and beats all three nulls under \textbf{global} and \textbf{pergroup}
  (p\_holm = 0.006). Under \textbf{rawglobal} it beats only N3 (p\_holm
  = 0.006; vs N1 p = 0.275, vs N2 p = 0.129). Laterality is thus the
  most calibration-robust property, but not significant against every
  null in every scheme.
\end{itemize}

An earlier internal result of a 43-fold learning-specificity advantage
was traced to a numerically sensitive calibration and is
\textbf{retracted}; under the robust schemes the ratio is 1.6--3.0 with
no significant difference between conditions.

\subsection{An innate head start and a different evolved sensory
strategy}\label{an-innate-head-start-and-a-different-evolved-sensory-strategy}

Pre-registered under \textbf{global} and \textbf{pergroup}; repeated
post hoc under \textbf{distmatch}.

{
\begin{longtable}[]{@{}
  >{\raggedright\arraybackslash}p{(\linewidth - 6\tabcolsep) * \real{0.2500}}
  >{\raggedright\arraybackslash}p{(\linewidth - 6\tabcolsep) * \real{0.2500}}
  >{\raggedright\arraybackslash}p{(\linewidth - 6\tabcolsep) * \real{0.2500}}
  >{\raggedright\arraybackslash}p{(\linewidth - 6\tabcolsep) * \real{0.2500}}@{}}
\toprule\noalign{}
\begin{minipage}[b]{\linewidth}\raggedright
endpoint
\end{minipage} & \begin{minipage}[b]{\linewidth}\raggedright
global
\end{minipage} & \begin{minipage}[b]{\linewidth}\raggedright
pergroup
\end{minipage} & \begin{minipage}[b]{\linewidth}\raggedright
distmatch (post hoc)
\end{minipage} \\
\midrule\noalign{}
\endhead
\bottomrule\noalign{}
\endlastfoot
generation 0 & A \textgreater{} N1, N2, N3 (10/10, 10/10, 10/10),
p\_holm 0.006 & A \textgreater{} all (9/10, 10/10, 10/10), p\_holm
0.006--0.008 & A \textgreater{} all (10/10, 9/10, 10/10), p\_holm
0.006 \\
olfaction ablation & A −0.03 vs +0.31 / +0.34 / +0.30; p\_holm
0.006--0.014 & A −0.08 vs +0.16 / +0.18 / +0.22; p\_holm 0.006--0.008 &
A −0.10 vs +0.25 / +0.32 / +0.28; p\_holm 0.006--0.008 \\
vision ablation & A +0.30 vs +0.13 / +0.14 / +0.07; p\_holm 0.06--0.11 &
A +0.33 vs +0.09 / +0.13 / +0.17; p\_holm 0.041 & A +0.47 vs +0.17 /
+0.13 / +0.05; p\_holm 0.012 \\
generation 275 & no difference vs N1, N2 (p\_holm 1.0); vs N3 p\_holm
0.059 & no difference vs N1, N2; vs N3 p\_holm 0.111 & no difference vs
N1, N2; vs N3 p\_holm 0.252 \\
AUC (primary) & A \textgreater{} N1, N3 (p\_holm 0.006); N2 n.s. & A
\textgreater{} N3 only (p\_holm 0.012) & A \textgreater{} N3 only
(p\_holm 0.012) \\
\end{longtable}
}

Median generation-0 differences are +0.16 to +0.37 fitness units. The
head start and the sensory-strategy divergence replicate in all three
calibrations, including the one that matches the full activity
distribution, so they are not explained by marginal activity statistics.
The cumulative endpoint is not robust: the degree-preserving null N2 is
never significantly worse on AUC or final fitness.

\subsection{No evidence that the nulls' olfactory reliance is
associative learning (post
hoc)}\label{no-evidence-that-the-nulls-olfactory-reliance-is-associative-learning-post-hoc}

On generation-250 population samples the reversal × plasticity
interaction is not significant in any condition or calibration (p\_holm
0.39--1.0), and extra toxic bites after an odor reversal do not differ.
A plasticity benefit of ≈ 0.09--0.30 exists in all conditions but is not
odor-specific. We therefore find \textbf{no evidence} that the nulls'
olfaction dependence is mediated by odor-specific associative learning;
with 10 seeds and the observed confidence intervals a smaller
interaction cannot be excluded.

Transfer: with a wider odor plume, connectome population samples
outperform N1 and N3 under \textbf{global} (p\_holm = 0.029); under
\textbf{distmatch} no transfer difference reaches significance, and
higher toxicity shows none in any scheme.

\subsection{The head start is amplified by, but not reducible to, the
engineered reflex (post
hoc)}\label{the-head-start-is-amplified-by-but-not-reducible-to-the-engineered-reflex-post-hoc}

Removing the innate feeding read-out reduces the generation-0 advantage
from +0.15\ldots+0.31 to +0.04\ldots+0.08, still significant in all
schemes (7--10/10 seeds, p 0.006--0.029). Blinding alone leaves it near
its original size (+0.13\ldots+0.30). Removing both vision and the
read-out leaves +0.03\ldots+0.07 (8--10/10 seeds, p ≤ 0.029). The
random-walk control, with all read-outs zeroed, gives a difference of
exactly 0.000 (0/10 seeds) in every scheme, confirming the differences
arise through the behavioural pathway and not through the evaluation
itself.

\subsection{Convergence with the degree-preserving null is not a ceiling
artifact (post
hoc)}\label{convergence-with-the-degree-preserving-null-is-not-a-ceiling-artifact-post-hoc}

Generation-250 population samples re-evaluated in harder worlds without
further evolution: at predator speed 0.18 the connectome still beats N1
and N3 (global +0.228 vs N1, 9/10, p = 0.010; distmatch +0.095 vs N1,
8/10, p = 0.027 and +0.342 vs N3, 8/10, p = 0.027) but is never
significantly different from N2 at any difficulty or in any scheme. At
the most extreme toxicity level the connectome advantage disappears in
all schemes.

\subsection{Negative result: neuron-level central
complex}\label{negative-result-neuron-level-central-complex}

A module of 1,041 real central-complex neurons (EPG, PEG, PEN, Delta7,
PFN, hΔB, FC2, PFL, plus GLNO/LNO/ER interface neurons with measured
connectivity) reproduces heading tracking of a rotating visual cue
(phase-tracking ratio 1.003; 93 \% of EPG variance on a two-dimensional
ring). Across ≈ 1,500 pathway-gain settings from random search and an
evolution strategy we found none that sustains heading in darkness,
integrates angular velocity, or supports decodable path integration, for
the connectome and the shuffled control alike. We report this as a
limitation of our rate-model formulation, not as a statement about the
connectome.

\begin{figure}
\centering
\includegraphics[width=1\linewidth,height=\textheight,keepaspectratio,alt={Common-garden fitness over 300 generations in paper-v1, per calibration scheme (median and interquartile range over 10 seeds).}]{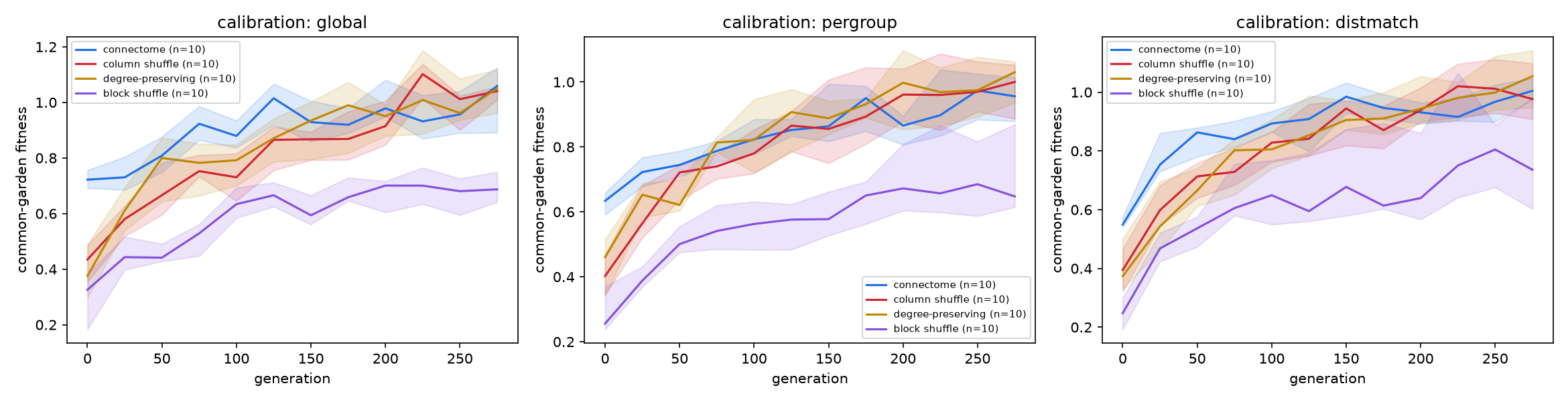}
\caption{Common-garden fitness over 300 generations in paper-v1, per
calibration scheme (median and interquartile range over 10
seeds).}\label{fig:v1-1}
\end{figure}

\begin{figure}
\centering
\includegraphics[width=1\linewidth,height=\textheight,keepaspectratio,alt={Paper-v1 endpoints per seed: generation 0, AUC (primary), generation 275, and ablation costs.}]{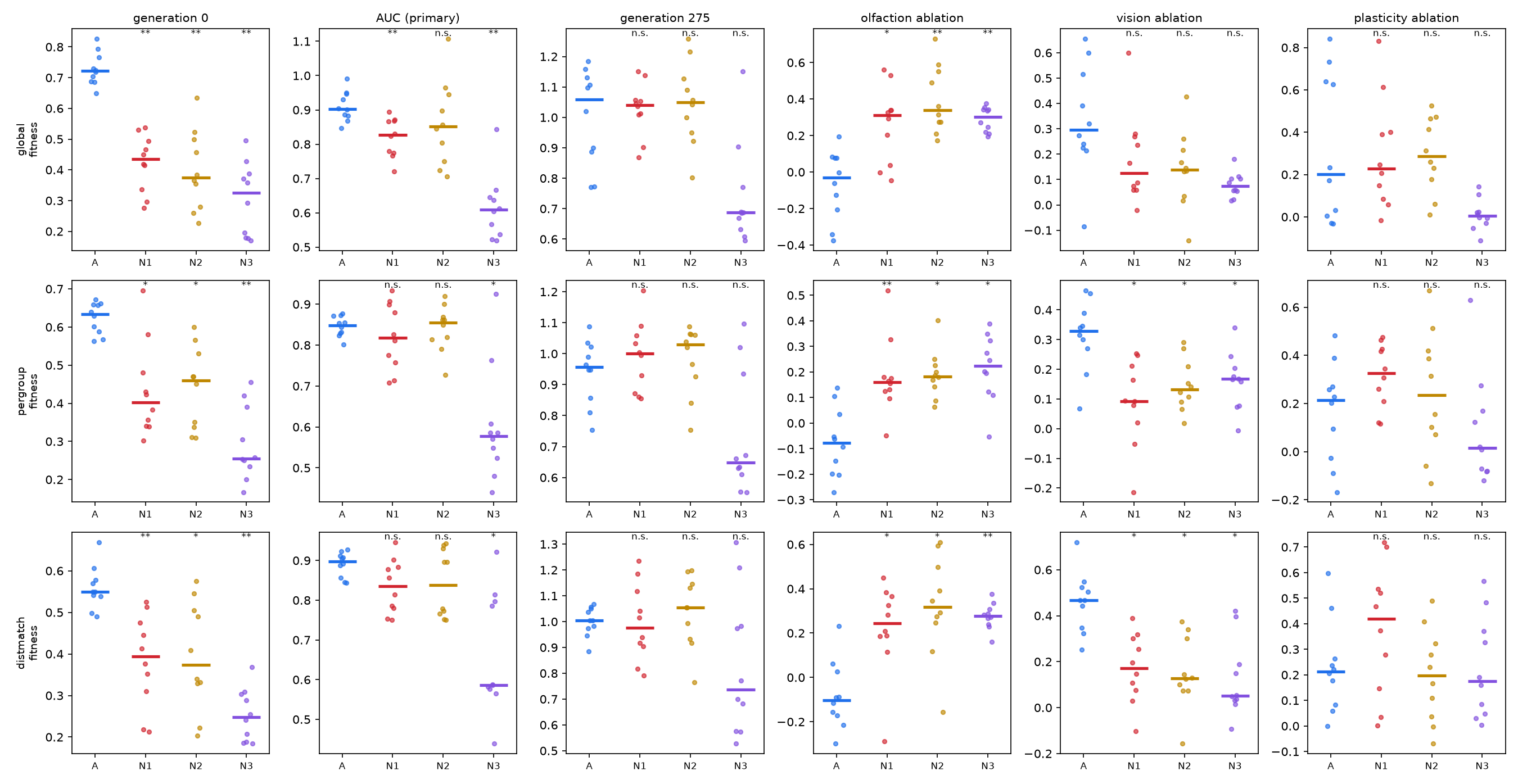}
\caption{Paper-v1 endpoints per seed: generation 0, AUC (primary),
generation 275, and ablation costs.}\label{fig:v1-2}
\end{figure}

\begin{figure}
\centering
\includegraphics[width=1\linewidth,height=\textheight,keepaspectratio,alt={Circuit metrics before evolution under three calibration schemes.}]{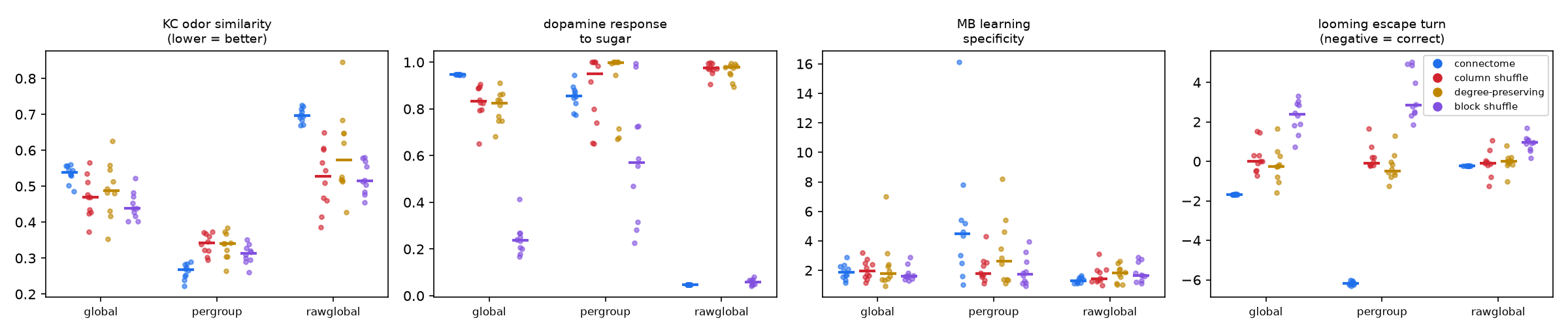}
\caption{Circuit metrics before evolution under three calibration
schemes.}\label{fig:v1-3}
\end{figure}

\begin{figure}
\centering
\includegraphics[width=1\linewidth,height=\textheight,keepaspectratio,alt={Generation-250 samples of paper-v1 tested in harder worlds without further evolution.}]{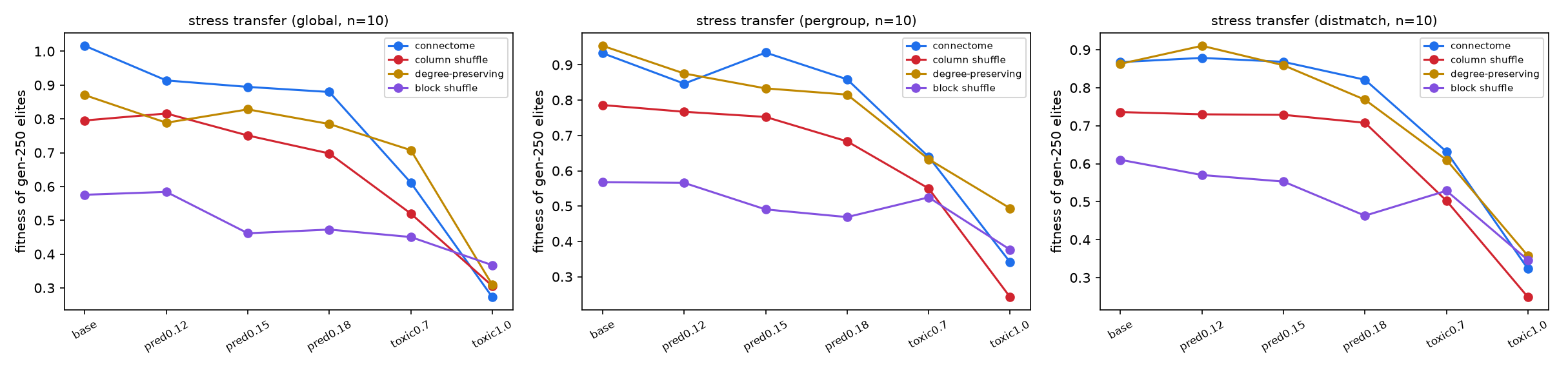}
\caption{Generation-250 samples of paper-v1 tested in harder worlds
without further evolution.}\label{fig:v1-4}
\end{figure}

\section{Use of AI tools}\label{app-ai}

The simulation engines, analysis and figure scripts were written with an
AI coding assistant (Claude Code, Anthropic), and the manuscript was
drafted with the same assistant from the author's instructions, the
stored result files and the dated research log. Two further language
models (Gemini, Google; GPT, OpenAI) were used to critique drafts and
analyses; their comments are referred to in the text as AI-assisted
reviews, not peer review, and the analyses added in response are
labelled post hoc. The mutational-neighbourhood assay, the turning-noise
control, the equivalence analysis and the interior-necessity assay were
added in response to them, and a question raised in one of them led to
the discovery of the noise error (Appendix \ref{app-prereg}). Every
number reported here is regenerated from the stored run outputs by
scripts in the repository (the data availability statement). The author
designed and directed the study, made every decision recorded in the
research log, and is responsible for the final text.

\end{document}